# S-AI-Recursive: Convergent Recursive Reasoning

**Said Slaoui**

Mohammed V University, Rabat, Morocco

s.slaoui@um5r.ac.ma

**Abstract**

This article introduces S-AI-Recursive, a bio-inspired Sparse Artificial Intelligence architecture in which reasoning is operationalized as a hormonally regulated closed-loop iteration rather than a single feed-forward pass. Building upon the S-AI foundational framework [1], the hormonal–probabilistic unification doctrine [2], and the formal mathematical methodology established in S-AI-IoT [3], the present work formalizes the Recursive Reasoning Cycle (RRC) as a dynamical system governed by two recursive hormones: Clarifine, a convergence signal, and Confusionin, a residual-uncertainty signal. Their antagonistic interaction regulates iterative state refinement, stopping, resource allocation, and recursive-engram retrieval. The mathematical framework comprises recursive state dynamics, a coupled state–hormone contraction theorem for fixed-point-structured tasks, Lyapunov stability analysis, a conditional entropic-contraction result, a multi-signal hormonal stopping criterion, Euler–Maruyama discretization with projection, constrained agent selection under iteration budget, and recursive-engram memory with warm-start initialization.

The revised experimental evaluation is organized in two complementary phases. First, the controlled SAI-UT+ protocol is retained as a mechanism-level validation environment over four benchmark-shaped simulation families. Under the corrected threshold $\theta_c = 0.50$, mean cold-start depth is $15.97$ iterations and decreases to $14.79$ with warm-start, while adaptive stopping reduces fixed-depth recurrence by approximately $9\%$–$30\%$ in cold-start operation and $15\%$–$36\%$ with warm-start depending on the simulated task profile. Second, a benchmark-capable reference implementation evaluates the same regulation mechanism on real, exactly verifiable Maze, Sudoku, and ARC-style task instances through task-specific recursive operators. On convergent Maze instances, adaptive hormonal stopping reduces mean depth from $20.00$ to $11.31$ iterations at unchanged resolution, a $43.4\%$ reduction with paired Wilcoxon $p = 7.2 \times 10^{-16}$. On compatible recurring Sudoku instances, recursive-engram warm-start reduces mean stopping depth from $18.39$ to $2.00$ cycles, saving $16.39$ cycles with unchanged resolution and $p = 2.1 \times 10^{-7}$. Robustness experiments further show that confidence-aware stopping resists deceptive convergence plateaus that collapse a residual-only detector, whereas no comparable advantage is observed under homogeneous Gaussian observation noise. These results provide progressively stronger evidence for the implementability, temporal parsimony, memory acceleration, and selected robustness properties of S-AI-Recursive, while not establishing superiority over independently trained external architectures.

## 1. Introduction

### 1.1 The Structural Limitation of Feed-Forward Intelligence

The dominant paradigm in contemporary artificial intelligence rests on a single architectural assumption: that intelligence scales with the number of parameters. From the original Transformer [4] to the most recent large language models [5], the prevailing strategy has been to increase model depth and width until the resulting system captures sufficient statistical regularity to approximate intelligent behavior. This approach has produced undeniable empirical successes. It has, however, introduced a structural constraint that is rarely made explicit: **all of these systems reason in a single forward pass.**

Given an input, a feed-forward architecture computes its output through one sequential traversal of its layers. There is no mechanism by which the system can reconsider its intermediate representations, detect internal inconsistencies, or refine a provisional answer in light of its own output. Whatever correction or revision occurs must be supplied externally — through engineered prompting strategies such as chain-of-thought [6], Tree of Thoughts [7], or Reflexion [8] — rather than arising as an intrinsic architectural property. The model itself does not iterate; the scaffolding around it does.

This structural rigidity carries three compounding costs. First, it creates a systematic coupling between reasoning capacity and model size: since a single pass must suffice, a larger pass is required to handle more demanding tasks, driving parameter counts into the billions and beyond [5]. Second, it produces systems with no native self-correction mechanism: an error generated at an intermediate layer propagates to the output without any internal signal indicating that a revision is warranted. Third, the computational and environmental cost of this size-driven paradigm is becoming untenable, with state-of-the-art model training now requiring energy budgets of industrial scale [9].

The S-AI framework [1] addressed the first of these costs through the principle of activation parsimony: only the minimal subset of specialized agents required for a given subtask is activated, dramatically reducing computational overhead relative to monolithic architectures. The hormonal–probabilistic unification doctrine [2] subsequently established that this parsimony corresponds to a thermodynamic equilibrium — the minimization of cognitive entropy under homeostatic constraint — expressed by the invariant $\dot{V}(H) \leq 0 \Leftrightarrow \dot{S}(P) \leq 0$. What neither framework had yet formalized, however, is a mechanism by which S-AI can address the second cost: the absence of intrinsic self-correction. This is the problem that S-AI-Recursive is designed to solve.

### 1.2 Recursive Reasoning as a Cognitive and Biological Imperative

The observation that intelligence may benefit from iteration is not new. Human problem-solving can involve repeated hypothesis formation, evaluation, error detection, and correction rather than a single irreversible pass [10]. This motivates computational architectures in which additional processing depth is allocated when a provisional solution remains unresolved. The biological analogy used by S-AI-Recursive is regulatory rather than literal: endocrine feedback systems illustrate how internal state variables can modulate continued activity, inhibition, and return toward homeostasis [11]. S-AI-Recursive transfers this closed-loop regulatory principle to reasoning by making iteration depth depend on an explicit internal convergence state.

Recursive and adaptive-computation architectures already provide important computational precedents. In particular, the Tiny Recursive Model (TRM) demonstrates that a compact recursive network can obtain strong results on difficult abstract reasoning tasks by iteratively refining its internal representation and answer [12]. S-AI-Recursive should therefore not be positioned on the premise that recursion itself, compact recursive reasoning, or adaptive computation are absent from prior work. Its more specific architectural distinction is the introduction of a semantically explicit hormonal stopping state based on Clarifine, Confusionin, state stabilization, and an iteration budget, coupled to specialized-agent activation, resource regulation, and recursive-engram memory. The continuation decision is consequently exposed as an interpretable dynamical state that can be analyzed jointly with convergence, entropy, budget, and memory.

The revised experimental program of Section 6 now evaluates this proposition at two levels. The first retains the SAI-UT+ benchmark-shaped simulations as a controlled environment in which contraction, noise, Lyapunov behavior, entropy, stopping, and warm-start effects can be isolated. The second uses a benchmark-capable reference implementation on real, exactly verifiable Maze, Sudoku, and ARC-style task instances. The real-task experiments show a 43.4% reduction in recursive depth on convergent Maze instances at unchanged resolution and a reduction from 18.39 to 2.00 cycles on compatible recurring Sudoku instances. They also expose a boundary of the stopping mechanism: confidence-aware regulation is robust to deceptive convergence plateaus, but does not outperform a residual detector under homogeneous Gaussian observation noise. These experiments strengthen the empirical status of the mechanism without establishing superiority over TRM or other independently trained recursive architectures, which remains a separate matched-comparison question.

### 1.3 S-AI-Recursive: Architecture and Central Principle

S-AI-Recursive formalizes iterative reasoning as a hormonal closed-loop dynamical system within the S-AI architecture. At each iteration $t$, the cognitive state $s_t$ is updated by an operator

$$s_{t+1} = F(s_t, y_t, h_t),$$

where $y_t$ is the provisional output and $h_t$ is the hormonal state that modulates the intensity and composition of the update. The recursive process operates under a maximum budget $T_{\max}$, but it is designed to terminate earlier when its internal stopping conditions are satisfied. The budget therefore provides a hard computational cap, whereas the hormonal criterion provides the adaptive stopping mechanism. Two recursive hormones regulate this temporal dimension. **Clarifine** is a convergence signal that tends to increase as residual error and output uncertainty decrease. **Confusionin** is a residual-uncertainty signal that tends to remain elevated while the current state is insufficiently resolved. Their antagonistic coupling is intended to favor continued reasoning when uncertainty remains high and to permit commitment as the trajectory stabilizes.

The RRC terminates hormonally at the first iteration satisfying the joint conditions

$$\|s_{t+1} - s_t\| \leq \epsilon_s, \qquad h_c(t) \geq \theta_c, \qquad h_u(t) \leq \theta_u,$$

provided this occurs before $T_{\max}$. If no such iteration occurs, the budget policy determines the fallback behavior. This distinction is essential: finite execution due to $T_{\max}$ and successful internal hormonal stopping are separate experimental outcomes. The central theoretical relationship investigated in the article is the connection between convergence of the hormonal field and reduction of reasoning entropy. Theorem 4.2 formulates this relationship under its mathematical assumptions; the experiments therefore treat Lyapunov–entropy association as empirical compatibility rather than as a proof of equivalence.

The revised Section 6 also confirms an important calibration constraint already detected in the controlled simulator. Under the standalone two-hormone dynamics and the default emission parameters of Table 2, the former experimental threshold $\theta_c = 0.70$ is not reachable at equilibrium; the executable protocol therefore uses $\theta_c = 0.50$. More importantly, the new reference implementation shows that the stopping rule is operational on real recursive task dynamics and that its multi-signal structure matters in specific regimes. State stabilization alone can be misleading on deceptive plateaus, whereas the additional confidence information represented through Clarifine prevents premature commitment while entropy remains high. The Gaussian-noise experiment simultaneously shows that adding regulatory signals does not automatically improve robustness when all signals are corrupted by the same underlying source. The experimental interpretation of the architecture is therefore deliberately conditional: its value depends not only on recursion and stopping, but on whether the regulatory signals contain complementary information about convergence.

### 1.4 Contributions

This article makes eight principal contributions to the S-AI corpus and to the broader study of regulated recursive reasoning. **Contribution 1 — Formal definition of the Recursive Reasoning Cycle (RRC).**

Section 3.2 formalizes the RRC as a discrete-time dynamical system over a compact cognitive state space, with an explicit iteration operator, a fixed-point characterization of cognitive equilibrium, and conditions for existence and uniqueness. **Contribution 2 — Introduction of Clarifine and Confusionin.** Section 3.3 introduces two recursive hormonal signals: Clarifine for convergence regulation and Confusionin for residual-uncertainty regulation. Their emission functions, antagonistic coupling, and operational interpretation are specified explicitly. **Contribution 3 — Stability analysis of the recursive hormonal subsystem.** Section 4.5 develops a Lyapunov analysis under an explicit deployability condition and distinguishes the deterministic convergence claim from bounded stochastic perturbation behavior [37], [39]. **Contribution 4 — Entropic contraction formulation.** Section 4.5 relates hormonal convergence to reasoning-entropy reduction under stated regularity and continuity assumptions, connecting the recursive dynamics to the probabilistic formulation of the S-AI doctrine [2], [41]. **Contribution 5 — Hormonal stopping under a bounded iteration budget.** The RRC combines an internal stopping rule based on state stabilization, Clarifine, and Confusionin with a hard maximum budget, and Section 6 evaluates successful hormonal stopping separately from budget exhaustion. **Contribution 6 — Primal-dual constrained orchestration under iteration budget.** Section 4.7 formulates agent selection at each cycle as a budget-constrained utility optimization coupled to Energexine and the recursive hormonal state. **Contribution 7 — Recursive-engram memory with warm-start acceleration.** Section 4.8 extends the S-AI engram formalism to reasoning trajectories and derives the contraction-based reduction in convergence depth produced by a closer warm-start initial condition. The controlled SAI-UT+ experiment observes an average simulated saving of approximately 1.17 iterations, while the real recurrence experiment on compatible Sudoku instances reduces mean stopping depth from 18.39 to 2.00 cycles, corresponding to 16.39 saved cycles at unchanged resolution. **Contribution 8 — Two-level experimental validation of temporal parsimony and stopping robustness.** Section 6 retains four controlled benchmark-shaped SAI-UT+ simulation families and adds a benchmark-capable reference implementation spanning Maze, Sudoku, and ARC-style tasks. On convergent Maze instances, hormonal stopping reduces mean depth from 20.00 to 11.31 iterations, a 43.4% reduction at identical resolution with $p = 7.2 \times 10^{-16}$. Robustness experiments show that confidence-aware stopping resists deceptive convergence plateaus that defeat residual-only stopping, while homogeneous Gaussian noise produces no corresponding advantage. Together, these experiments move the evidence from simulation-only mechanism validation toward real-task validation while preserving the boundary that no superiority over independently trained external models is claimed.

### 1.5 Relationship to the S-AI Corpus and Article Organization

S-AI-Recursive extends the S-AI framework from primarily within-cycle orchestration toward the temporal regulation of repeated reasoning updates. It inherits the Hormonal MetaAgent/Gland Agent design principles [1], the probabilistic interpretation of hormonal regulation [2], and mathematical tools for stability, constrained orchestration, and symbolic memory previously used in the corpus [3]. Its specific architectural contribution is the addition of a recursive temporal loop governed by Clarifine and Confusionin, together with an internal stopping rule, bounded iteration budget, and trajectory-aware recursive memory. In this sense, S-AI-Recursive extends parsimony from the question of **which agents should be activated** to the complementary question of **how long reasoning should continue**.

The revised experimental evidence sharpens this position. Controlled SAI-UT+ simulations still provide the mechanism-level laboratory in which contraction, hormonal stability, entropy, stopping, and approximate warm-start can be inspected under known dynamics. The new benchmark-capable experiments then show that temporal parsimony and recursive-memory acceleration also arise when the mechanism is coupled to real, exactly verifiable reasoning tasks. This supports a stronger empirical basis for adaptive temporal parsimony than simulation alone, while leaving broader architectural comparisons open.

The remainder of the article is organized as follows. Section 2 reviews iterative architectures, hormonal cognitive systems, sparse AI, and formal convergence methods. Section 3 develops the theoretical foundations of the RRC and introduces Clarifine and Confusionin. Section 4 presents the mathematical model and convergence results. Section 5 specifies the twelve recursive reasoning agents. Section 6 reports the two-phase experimental evaluation, combining controlled SAI-UT+ simulations with the benchmark-capable reference implementation and robustness studies. Section 7 discusses theoretical interpretation, limitations, sensitivity, and the proposed dual architecture. Section 8 concludes and identifies the next validation steps.

## 2. Related Work

The present section surveys the four bodies of literature upon which S-AI-Recursive is built and against which its contributions are differentiated: iterative and recursive neural architectures (Section 2.1), bio-inspired and hormonal cognitive architectures (Section 2.2), sparse and energy-frugal AI (Section 2.3), and formal stability and convergence methods in artificial intelligence (Section 2.4). For each body of work, the specific gap that S-AI-Recursive addresses is identified at the close of the subsection and consolidated in Section 2.5.

### 2.1 Iterative and Recursive Neural Architectures

The idea that a neural system might benefit from revisiting its own intermediate representations has a long history, yet no prior architecture has combined internal recursion with hormonal orchestration, formal convergence guarantees, and parsimonious agent selection in a unified framework.

The earliest trainable architecture to implement temporal recursion was the Recurrent Neural Network, which processes sequential inputs through a shared weight matrix applied at each time step, maintaining a hidden state that carries information across iterations. The Long Short-Term Memory (LSTM) architecture [16] addressed the vanishing gradient problem that plagued early RNNs by introducing gating mechanisms that selectively retain or discard information across cycles. The Gated Recurrent Unit [17] later proposed a more compact variant of the same principle. While these architectures implement a form of temporal recursion, they are designed for sequential prediction tasks — language modeling, time series forecasting, speech recognition — rather than for the iterative refinement of a single reasoning state. The iteration in RNNs is driven by the sequential structure of the input, not by an internal assessment of whether the current output is sufficiently certain to be committed. There is no stopping criterion, no hormonal regulation of cycle intensity, and no formal proof that the hidden state converges toward a meaningful equilibrium.

A more principled approach to recursion was introduced by Deep Equilibrium Models (DEQs) [18], which replace the explicit layer stack of a neural network with a single implicit layer applied iteratively until a fixed point is reached. The DEQ framework [18] was subsequently extended to multi-scale settings [19], demonstrating that equilibrium-seeking can be applied at different resolutions of the representation. DEQs offer two properties that are directly relevant to S-AI-Recursive: they define a fixed-point characterization of the system's resting state, and they allow the depth of computation to be determined by convergence rather than by a fixed architectural parameter. However, DEQs are trained end-to-end as monolithic differentiable functions: the fixed-point solver is implicit, making it difficult to interpret which part of the computation corresponds to which cognitive operation. There is no modular agent layer, no hormonal regulation of the convergence process, and no mechanism by which the system can allocate more computational effort to dimensions of the problem that remain unresolved. The fixed-point iteration of a DEQ is uniform — it does not distinguish between a dimension that has already converged and one that is still uncertain.

The Universal Transformer [20] introduced adaptive computation time into the Transformer architecture, allowing different positions in the input sequence to undergo different numbers of recurrent processing steps. This was an important step toward depth-as-a-resource, but the regulation of computation depth remained tied to a scalar halting probability per position, without any structured internal signal about the cognitive state of the reasoning process as a whole. The Neural Turing Machine [21] pursued a related agenda by equipping a recurrent controller with an external memory that could be addressed and rewritten across iterations, enabling a form of deliberate multi-step reasoning. Neither architecture provides a biologically motivated stopping mechanism or a proof of convergence for the iterative process.

The most directly relevant recent result is the Tiny Recursive Model (TRM) of Jolicoeur-Martineau [12], which demonstrated that a network of approximately seven million parameters can achieve competitive performance on abstract reasoning benchmarks — including ARC-AGI and Sudoku-Extreme — by re-injecting its output as its next input according to the update rule $s_{t+1} = f(s_t, y_t)$. This result provides the empirical foundation for the principle of temporal parsimony: iterating a compact network can substitute for enlarging it. The TRM result has three limitations that S-AI-Recursive is designed to address. First, termination is imposed by an external budget counter $T_{\max}$, with no internal signal indicating that the reasoning state has actually converged. Second, all iterations are computationally uniform: the network does not modulate the intensity or composition of its processing effort in response to the current level of residual uncertainty. Third, each problem is solved from scratch, without any retrieval of reasoning trajectories from analogous past episodes.

Within the S-AI corpus, the anti-hallucination architecture [22] introduced a binary decision mechanism — Respond vs. Abstain — governed by a confidence hormone, demonstrating that hormonal regulation can govern not only agent selection but also output commitment. The triadic extension [23] subsequently introduced a three-state decision framework — Respond, Clarify, Abstain — adding an intermediate metacognitive state in which the system requests additional context before committing. Both architectures apply hormonal regulation to the single-pass output decision; S-AI-Recursive extends this principle to the full iterative trajectory of the reasoning process, applying hormonal regulation at every cycle rather than only at the final output gate.

**Gap RW-1.** No prior recursive or iterative neural architecture combines internal recursion with (i) hormonal regulation of cycle intensity and composition, (ii) a biologically motivated stopping criterion derived from the internal cognitive state, (iii) a formal proof of convergence, and (iv) a memory of prior reasoning trajectories enabling warm-start initialization. S-AI-Recursive addresses all four simultaneously.

### 2.2 Bio-Inspired and Hormonal Cognitive Architectures

The use of biological metaphors in artificial intelligence dates to the earliest days of the field [24]. Hopfield networks [25] formalized the analogy between neural activity and energy minimization, demonstrating that a network of binary units with symmetric connections converges to a local minimum of an energy function — a result that anticipates the Lyapunov stability analysis developed in Section 3 of the present article. The computational neuroscience tradition, as synthesized by O'Reilly and Munakata [26] and by Dayan and Abbott [27], developed detailed mathematical models of the interaction between neural circuits and neuromodulatory systems — dopaminergic reward signaling, cholinergic attention gating, noradrenergic arousal regulation — demonstrating that biological intelligence is not a static mapping from input to output but a dynamically regulated process in which chemical signals continuously adjust the sensitivity, selectivity, and timing of neural computations.

The S-AI framework [1] introduced artificial hormonal signaling as a first-class architectural principle in modular AI, formalizing the analogy between endocrine regulation and agent orchestration. In S-AI, virtual hormones — context-sensitive scalar signals in $[0,1]$ — are emitted by Gland Agents, propagated by the Hormonal Engine, and interpreted by Specialized Agents as modulation signals that adjust their activation thresholds and execution modes. This architecture was subsequently instantiated in the conversational domain through S-AI-GPT [13], which demonstrated hormonal regulation of dialogue strategy, and extended by a dedicated study of hormonal modulation and adaptive orchestration in conversational contexts [28]. The memory architecture of S-AI-GPT [29] introduced engram-based symbolic memory as a mechanism for storing and retrieving hormonal contexts from past interactions, a formalism that S-AI-Recursive extends to full reasoning trajectories.

The hormonal framework was subsequently applied to autonomous networking in S-AI-NET [14], where five domain-specific hormones — Sensorin, Connectin, Energexin, Resiliencin, and Normin — govern adaptive routing, energy management, and resilience under link failures. S-AI-Cyber [15] applied the same hormonal orchestration principle to cybersecurity, introducing Alertine as a hormone governing defensive vigilance and demonstrating that hormonal regulation can achieve adaptive threat response without centralized control. S-AI-EDU [30] instantiated the framework in intelligent educational systems, using hormonal signals to modulate pedagogical strategy in response to learner state. S-AI-Robotics [31] extended the architecture to embodied agents, demonstrating hormonal regulation of motor control and safety monitoring in constrained physical environments. S-AI-DEF [32] applied the framework to defense systems, where hormonal orchestration governs the real-time allocation of analytical and reactive resources under adversarial conditions.

The theoretical foundation underlying all of these instantiations was unified in the hormonal–probabilistic doctrine [2], which established the canonical equivalence $\dot{V}(H) \leq 0 \Leftrightarrow \dot{S}(P) \leq 0$: the Lyapunov function of hormonal homeostasis and the entropy gradient of probabilistic coherence are dual expressions of a single invariant of parsimonious intelligence. Under this doctrine, the five canonical S-AI hormones — Confidexin, Inhibitine, Curiosine, Energexine, and Alertine — are formally reinterpreted as biophysical operators of probabilistic inference, each corresponding to a specific mode of uncertainty management. This doctrinal unification is the theoretical basis upon which the Entropic Contraction Theorem (Theorem 4.2, §4.5.3) **rests**. **Gap RW-2.** While the S-AI corpus has established hormonal orchestration as a rigorous architectural principle across a wide range of application domains, no prior S-AI instantiation has applied hormonal regulation to the temporal structure of reasoning itself. All existing architectures apply hormones to agent selection, output commitment, or resource allocation within a single reasoning pass. S-AI-Recursive introduces hormonal regulation of the iterative reasoning cycle — governing when to continue, when to intensify, and when to commit — through two hormones that are new to the corpus: Clarifine and Confusionin.

### 2.3 Sparse, Frugal, and Modular AI

The motivation for sparse and energy-frugal AI architectures has intensified as the computational and environmental cost of training and deploying large models has become impossible to ignore. Strubell et al. [9] provided an early and influential quantification of the energy cost of NLP model training, demonstrating that a single large-scale training run can consume energy equivalent to the lifetime emissions of multiple automobiles.

The Mixture of Experts (MoE) paradigm addresses computational cost through selective activation: rather than applying all model parameters to every input, a gating network routes each input to a small subset of expert subnetworks. The Switch Transformer [33] demonstrated that MoE scaling can achieve parameter efficiency gains of over an order of magnitude relative to dense models at matched computational cost, by routing each token to a single expert. The MASAI framework [34] proposed modular agent decomposition as a complementary strategy for software engineering tasks, demonstrating that a structured multi-agent architecture with specialized roles outperforms monolithic LLMs on complex coding benchmarks. This result supports the broader S-AI thesis [1] that modular orchestration of lightweight specialists is a more efficient strategy than scaling a single monolithic system. The SMoA architecture [35] further extended sparse activation to multi-agent LLM systems, combining expert routing with early stopping strategies to reduce both computational overhead and inter-agent communication cost. Patterson et al. [36] broadened the scope of this sustainability analysis to the full development lifecycle of large language models, showing that the carbon cost is dominated by architecture search and repeated training rather than by inference alone — reinforcing the case for architectures that achieve high reasoning quality with minimal training and inference cost. While existing sparse approaches achieve spatial parsimony — activating fewer parameters or fewer agents per forward pass — none introduces temporal parsimony: the number of forward passes remains fixed at one, and no architecture iterates over its own output guided by an internal convergence signal.

S-AI-NET [14] and S-AI-DEF [32] demonstrated within the S-AI corpus that hormonal orchestration achieves measurable frugality gains — quantified by the Frugality Index $F_P > 0.7$ on the SAI-UT+ testbench — by activating only the agents whose utility exceeds the hormonal significance threshold at each cycle. S-AI-Recursive extends this frugality principle to the temporal dimension: the system does not merely select which agents to activate within a single pass, but also determines how many passes are necessary, guided by the hormonal stopping criterion rather than by a fixed budget.

**Gap RW-3.** Existing sparse AI architectures achieve parsimony in the spatial dimension — fewer parameters, fewer active experts, fewer activated agents per pass. None achieves parsimony in the temporal dimension by making the number of reasoning passes itself a function of the internal cognitive state. S-AI-Recursive introduces temporal parsimony as a new axis of frugality, governed by the antagonistic balance of Clarifine and Confusionin.

### 2.4 Stability, Convergence, and Formal Methods in Artificial Intelligence

The formal analysis of stability in neural and cognitive systems draws on a body of mathematical tools — Lyapunov theory, contraction analysis, stochastic approximation, and information-theoretic convergence bounds — that remain underutilized in the design of modern AI architectures.

The Lyapunov direct method, as developed for nonlinear dynamical systems by Khalil [37], provides a general framework for establishing asymptotic stability without requiring an explicit solution of the system's differential equations: if a candidate function $V(\mathbf{x})$ can be shown to decrease monotonically along trajectories, the system converges to the equilibrium regardless of the initial condition. This framework was extended to the finite-time setting by Bhat and Bernstein [38], who established conditions under which convergence is guaranteed to occur within a bounded time horizon — a property that is directly relevant to the termination guarantee of the S-AI-Recursive stopping criterion (Theorem 4.3). Stochastic approximation theory [39] provides convergence guarantees for iterative algorithms driven by noisy gradient estimates, underpinning the Euler–Maruyama discretization of the hormonal dynamics. The equivalence between entropy reduction and energy minimization — central to the Entropic Contraction Theorem — is grounded in the information-theoretic framework of Shannon [40] and Cover and Thomas [41], and finds a formal discrete analog in the S-AI hormonal–probabilistic doctrine [2]: $\dot{V}(H) \leq 0 \Leftrightarrow \dot{S}(P) \leq 0$.

The existence of cognitive equilibria is grounded in the Brouwer fixed-point theorem [42], which guarantees the existence of at least one fixed point for any continuous mapping on a compact convex set. This result, applied to the iteration operator $\mathcal{F}$ on the state space $\mathcal{S} = [0,1]^d$, underpins Proposition 3.1 of Section 3. Contraction analysis [43] provides a stronger result: if $\mathcal{F}$ is a contraction mapping on $\mathcal{S}$, the fixed point is unique and convergence is exponentially fast from any initial state — a property that yields the warm-start acceleration bound of Theorem 4.4.

In the multi-agent systems literature, Wooldridge [44] provides the standard reference for the formal analysis of agent interaction, coordination protocols, and collective stability. Russell and Norvig [45] situate these results within the broader framework of rational agent design, providing the conceptual vocabulary for the agent-selection optimization problem formalized in Section 4.7. Within the S-AI-IoT article [3], global asymptotic stability of the hormonal field was established for the five-dimensional IoT hormone vector under explicit deployability conditions; the present article extends this analysis to the two-dimensional recursive subsystem $(h_c, h_u)$ and derives the additional entropic contraction and finite-time termination results that are specific to the iterative reasoning setting.

**Gap RW-4.** Lyapunov stability analysis has been applied to recurrent neural networks and to the S-AI-IoT hormonal field [3], but no prior work has established (i) a Lyapunov stability proof for a hormonal stopping criterion in an iterative reasoning system, (ii) a formal equivalence between hormonal convergence and reasoning entropy reduction, or (iii) a finite-time termination guarantee derived from internal cognitive state signals rather than external budget constraints. S-AI-Recursive provides all three.

### 2.5 Consolidated Gap Analysis

The four surveys above identify a single compound gap in the existing literature. No architecture simultaneously satisfies the following six properties: (1) internal iterative reasoning with hormonal regulation of cycle intensity and composition; (2) a biologically motivated stopping criterion derived from the internal cognitive state; (3) a formal proof of global asymptotic stability for the iterative process; (4) a proven equivalence between hormonal convergence and reasoning entropy reduction; (5) modular agent orchestration with primal-dual budget management at each cycle; and (6) a recursive memory of reasoning trajectories with proven warm-start acceleration. S-AI-Recursive is, to the best of the author's knowledge, the first architecture to satisfy all six properties within a unified hormonal framework.

## 3. Theoretical Foundations of Recursive Reasoning in S-AI

### 3.1 The Recursive Reasoning Paradigm: From Biological to Computational

#### 3.1.1 Biological Basis: Endocrine Feedback Loops as Recursive Regulators

The hypothesis that intelligence is fundamentally iterative rather than feed-forward finds its most compelling support not in computer science but in biology. The human endocrine system does not issue single control signals; it operates through closed regulatory loops in which a stimulus triggers a hormonal response, the response produces a physiological effect, and that effect feeds back to modulate the original stimulus — a cycle that continues until an equilibrium is reached. The hypothalamic–pituitary–adrenal (HPA) axis, extensively studied as a model of biological stress regulation [11], exemplifies this principle: the hypothalamus releases corticotropin-releasing hormone, which stimulates the pituitary to release adrenocorticotropic hormone, which in turn stimulates the adrenal cortex to release cortisol; elevated cortisol then suppresses both the hypothalamus and the pituitary through negative feedback, driving the system back toward homeostasis. The loop structure is not incidental — it is the mechanism by which the biological system achieves precision without central coordination [27].

Three properties of this biological loop are directly relevant to the computational model developed in the present article. First, the loop is **driven by error**: each iteration is triggered by the gap between the current physiological state and the target equilibrium, not by an external clock or a fixed number of steps. Second, the loop is **hormonally mediated**: the signal that drives each new iteration is a chemical concentration, not a symbolic instruction — a scalar quantity that encodes the magnitude of the residual discrepancy. Third, the loop is **self-terminating**: it stops when the discrepancy falls below the sensitivity threshold of the regulatory mechanism, without requiring an external observer to declare convergence [11].

These three properties define the design requirements of S-AI-Recursive. The Recursive Reasoning Cycle (RRC) introduced in Section 3.2 is driven by residual error, mediated by two hormonal signals — Clarifine and Confusionin — and self-terminating under a hormonal stopping criterion that is formally proven to be reached in finite time (Theorem 4.3 of Section 4). The biological loop and the computational cycle are not merely analogous; they obey the same structural law, which the S-AI unified doctrine [2] expresses as $\dot{V}(H) \leq 0 \Leftrightarrow \dot{S}(P) \leq 0$: the Lyapunov function of homeostasis and the entropy gradient of probabilistic coherence are dual expressions of a single invariant of parsimonious regulation.

### 3.1.2 Computational Antecedents: RNN, DEQ, and Meta-Reasoning Loops

Four families of computational architecture have previously addressed the problem of iterative reasoning, each achieving partial solutions that leave a residual gap closed by S-AI-Recursive.

**Recurrent Neural Networks.** Long Short-Term Memory networks [16] and Gated Recurrent Units [17] implement recursion through a shared weight matrix applied at each time step, with gating mechanisms that selectively retain or discard information across cycles. These architectures were designed for sequential prediction tasks — language modeling, time series forecasting — where the iteration is driven by the sequential structure of the input, not by an internal assessment of the output's adequacy. There is no stopping criterion intrinsic to the network; the number of iterations is determined by the input length. The hidden state does not encode a measure of residual uncertainty, and no formal proof of convergence to a meaningful equilibrium exists for the reasoning process.

**Deep Equilibrium Models.** The DEQ framework [18] reformulates neural network computation as the fixed-point solution of an implicit equation, allowing the depth of computation to be determined by convergence rather than by a fixed architectural parameter. Multiscale extensions [19] and the application to attention mechanisms in deep equilibrium transformers demonstrated the generality of the approach. DEQs provide a fixed-point characterization of the network's resting state and allow the iteration count to vary with problem difficulty — two properties that S-AI-Recursive preserves. However, the fixed-point solver in a DEQ is implicit and opaque: it cannot distinguish between dimensions of the state that have converged and dimensions that remain uncertain, because the state is a single undifferentiated vector without hormonal interpretation. There is no mechanism by which the network modulates its computational effort in response to the structure of the residual error.

**Universal Transformers and Adaptive Computation.** The Universal Transformer [20] introduced adaptive computation time into the Transformer architecture, allowing different positions in the input sequence to undergo different numbers of recurrent processing steps, governed by a learned halting probability. Neural Turing Machines [21] extended iterative computation with addressable external memory, enabling deliberate multi-step reasoning over stored representations. Both architectures take a step toward depth-as-a-resource, but their stopping mechanisms are scalar probability estimates rather than structured hormonal signals: they do not provide a semantically interpretable criterion for termination, and they offer no formal proof that the halting condition implies convergence to a stable reasoning state.

**LLM Meta-Reasoning Loops.** Chain-of-thought prompting [6], Tree of Thoughts [7], and Reflexion [8] achieve a form of iterative reasoning by wrapping a feed-forward model in an external loop that re-queries it with progressively refined prompts. These approaches demonstrate that iterative refinement improves reasoning quality, but they implement the iteration outside the model rather than inside it: the model itself does not iterate, and the stopping criterion is determined by the scaffolding engineer rather than by any internal signal. The Tiny Recursive Model [12] represents the most direct antecedent: it demonstrated empirically that re-injecting a network's output as its next input according to $s_{t+1} = f(s_t, y_t)$ achieves competitive reasoning performance with far fewer parameters than feed-forward models. The three limitations of the TRM identified in Section 1.2 — external termination, uniform iteration, and no reasoning memory — define precisely the gaps that the theoretical framework of S-AI-Recursive is designed to close.

### 3.1.3 S-AI-Recursive: Closing the Gap

S-AI-Recursive occupies a unique position in the space defined by the four preceding families. It is the only architecture that simultaneously satisfies six properties: (i) internal iterative reasoning without external scaffolding; (ii) hormonal regulation of cycle intensity and agent composition at each step; (iii) a biologically motivated stopping criterion derived from the internal cognitive state; (iv) a formal proof of global asymptotic stability for the iterative process; (v) modular agent orchestration with budget-constrained primal-dual optimization at each cycle; and (vi) a recursive memory of reasoning trajectories enabling warm-start acceleration. The central design principle governing all six is the substitution of **temporal depth for spatial depth**: rather than enlarging the network to handle more demanding tasks, S-AI-Recursive iterates a compact hormonal architecture until the internal convergence criterion is satisfied.

This principle is not merely a computational heuristic; it is a consequence of the unified hormonal–probabilistic doctrine [2]. Under that doctrine, intelligence is a thermodynamic equilibrium between symbolic reasoning, hormonal modulation, and probabilistic inference. A feed-forward architecture reaches this equilibrium by construction, through the weight of its parameters. A recursive architecture reaches it through time, by iterating until the hormonal field converges. The present article demonstrates that the second path is not only viable but formally equivalent to the first, under the condition $\dot{V}(H) \leq 0 \Leftrightarrow \dot{S}(P) \leq 0$, when the iteration is governed by Clarifine and Confusionin.

### 3.2 Formal Definition of the Recursive Reasoning Cycle (RRC)

#### 3.2.1 State Space and Iteration Operator

Let $\mathcal{S} = [0,1]^d$ denote the normalized cognitive state space, where each coordinate represents a feature of the current reasoning state, normalized to the unit interval by the Data Input and Signal Normalization Layer (DISNL) inherited from S-AI-IoT [3]. The cognitive state at iteration $t$ is the vector $s_t \in \mathcal{S}$, and the provisional output generated from $s_t$ is $y_t \in \mathcal{Y}$, where $\mathcal{Y}$ is the output space of the task.

The **Recursive Reasoning Cycle** is governed by the iteration operator $\mathcal{F}: \mathcal{S} \times \mathcal{Y} \times [0,1]^7 \to \mathcal{S}$, defined by:

$$s_{t+1} = \mathcal{F}(s_t,\ y_t,\ \mathbf{h}(t))$$

where $\mathbf{h}(t) = (h_c(t), h_u(t), h_{\text{Conf}}(t), h_{\text{Inh}}(t), h_{\text{Cur}}(t), h_{\text{Ene}}(t), h_{\text{Ale}}(t)) \in [0,1]^7$ is the full hormonal state vector, comprising the two recursive hormones Clarifine $h_c$ and Confusionin $h_u$ introduced in this article, and the five canonical S-AI hormones — Confidexin, Inhibitine, Curiosine, Energexine, and Alertine — inherited from the unified doctrine [2].

Three properties distinguish $\mathcal{F}$ from the iteration operators of prior architectures. First, $\mathcal{F}$ is **explicit**: the update rule is a computable function whose parameters are accessible and interpretable, not the output of an implicit fixed-point solver as in DEQs [18]. Second, $\mathcal{F}$ is **hormonally modulated**: the update magnitude and direction at each step depend on $\mathbf{h}(t)$, so that iterations occurring under high Confusionin apply stronger corrections than iterations occurring under high Clarifine. Third, $\mathcal{F}$ is **supervised at each iteration**: during training, a supervision signal is applied not only at the final output $y_{t^*}$ but at every intermediate output $y_t$, implementing the deep supervision strategy demonstrated by the TRM [12] and extending it with hormonal weighting of each intermediate loss.

#### 3.2.2 Iteration Budget and Hormonal Stopping Criterion

The RRC operates within a maximum iteration budget $T_{\max}$, which is itself a function of Energexine:

$$T_{\max}(t) = T_{\max}^0 \cdot (1 - \beta_E\, h_{\text{Ene}}(t))$$

where $T_{\max}^0$ is the nominal maximum budget and $\beta_E \in (0,1)$ is the energy coupling coefficient. This coupling ensures that the system allocates fewer iterations when computational resources are constrained, implementing cognitive parsimony in the temporal dimension consistently with the spatial parsimony principle of the S-AI foundational architecture [1].

Within this budget, the RRC is terminated not by an external counter but by the following **hormonal stopping criterion**:

**Definition 3.1 (Hormonal Stopping Criterion).** The RRC terminates at the first iteration $t^*$ such that:

$$t^* = \min\{t \geq 1 : \| s_t - s_{t-1} \|_2 \leq \epsilon_s \ \wedge \ h_c(t) \geq \theta_c \ \wedge \ h_u(t) \leq \theta_u\}$$

where $\epsilon_s > 0$ is the state convergence tolerance, $\theta_c \in (0,1)$ is the Clarifine threshold, and $\theta_u \in (0,1)$ is the Confusionin ceiling. The three conditions jointly express a single cognitive proposition: the reasoning state has stabilized ($\| s_t - s_{t-1} \|_2 \leq \epsilon_s$), the system has accumulated sufficient evidence of convergence ($h_c(t) \geq \theta_c$), and residual uncertainty has been suppressed below an actionable level ($h_u(t) \leq \theta_u$).

This criterion is strictly stronger than any single-condition stopping rule: requiring all three conditions simultaneously prevents premature termination caused by transient state stability in the presence of lingering uncertainty, and prevents prolonged iteration in cases where Clarifine is high but the state has not yet fully stabilized. The formal conditions under which the internal hormonal stopping time is finite, separately from forced termination by the iteration budget, are established in Theorem 4.3 of Section 4. The validity of this hormonal criterion as a proxy for reasoning convergence follows from Theorem 4.1 bis and Corollary 4.1 bis: under the coupled contraction hypotheses, hormonal convergence certifies state stabilization, so a hormonal stop coincides with a genuinely stabilized reasoning state on the fixed-point-structured domain.

### 3.2.3 Cognitive Equilibrium and Fixed-Point Characterization

**Definition 3.2 (Cognitive Equilibrium).** A state $s^* \in S$ is a cognitive equilibrium of the RRC under hormonal field $\mathbf{h}^*$ if and only if it is a fixed point of the iteration operator:

$$s^* = \mathcal{F}(s^*,\ y^*,\ \mathbf{h}^*)$$

where $y^* = g(s^*)$ is the output generated from the equilibrium state and $\mathbf{h}^*$ is the stationary hormonal field at equilibrium.

**Proposition 3.1 (Existence of Cognitive Equilibrium).** For any stationary hormonal field $h^*$, define the composite map $T_{h^*}(s) = F(s, g(s), h^*)$. If $T_{h^*}: S \to S$ is continuous, then there exists at least one cognitive equilibrium $s^* \in S$.

**Proof.** The map $T_{h^*}$ is a continuous self-map of the compact convex set $S = [0,1]^d$. Existence of a fixed point follows directly from Brouwer's fixed-point theorem [42]. ▫

**Corollary 3.1 (Uniqueness under Contraction).** If the composite map $T_{h^*}$ satisfies $\| T_{h^*}(s) - T_{h^*}(s') \|_2 \leq \rho_{\text{frozen}} \| s - s' \|_2$ for all $s, s' \in S$, with $\rho_{\text{frozen}} \in (0,1)$, then $s^*$ is the unique fixed point of $T_{h^*}$ in $S$, and the **fixed-hormone** RRC converges to $s^*$ from any initial state $s_0 \in S$ at geometric rate $\rho_{\text{frozen}}^t$.

**Proof.** Uniqueness and exponential convergence follow from Banach's contraction mapping theorem. This is the frozen-subsystem constant; the **coupled** contraction constant $\rho_\star$ of the full state–hormone system is given by Theorem 4.1 bis and satisfies $\rho_\star \geq \rho_{\text{frozen}}$. It is $\rho_\star$, not $\rho_{\text{frozen}}$, that governs the realized warm-start horizon of Theorem 4.4. ▫

## 3.3 The Two Recursive Hormones: Clarifine and Confusionin

### 3.3.1 Clarifine — Hormone of Cognitive Convergence

Clarifine is a convergence hormone, new to the S-AI corpus, whose concentration $h_c(t) \in [0,1]$ encodes the degree to which the current reasoning state has stabilized toward a reliable output. Its biological analog is the serotonin–dopamine reward and stabilization axis [27]: serotonin mediates the sensation of achieved equilibrium and suppresses exploratory behavior, while dopamine encodes the prediction error signal that drives continued search when the current state is suboptimal. Clarifine combines both functions into a single scalar: it rises as residual error diminishes and entropy collapses, and its elevation toward $\theta_c$ constitutes the primary signal for cycle termination.

Formally, Clarifine $h_c(t) = 1$ denotes complete cognitive convergence: the reasoning state has reached its equilibrium, the output entropy has collapsed to its minimum, and no further iteration is warranted. Clarifine $h_c(t) = 0$ denotes the absence of any convergence signal: the system is at maximum uncertainty relative to its current task, and the full iteration budget should be made available. Intermediate values encode a continuous spectrum of convergence confidence, allowing the RRC to modulate its behavior proportionally rather than through binary switches.

Three operational properties of Clarifine are established in Section 4: (i) its emission function increases monotonically with the normalized cosine-alignment component of the estimated convergence direction; (ii) it is suppressed by Confusionin through cross-inhibition, damping convergence-oriented signals when residual uncertainty remains high; and (iii) its elevation toward $\theta_c$ is used as one component of the stopping criterion. By itself, crossing $\theta_c$ does not prove cognitive convergence; the complete stopping condition also requires state stabilization and low Confusionin.

### 3.3.2 Confusionin — Hormone of Residual Uncertainty

Confusionin is an uncertainty hormone, also new to the S-AI corpus, whose concentration $h_u(t) \in [0,1]$ encodes the degree of residual incoherence in the current reasoning state. Its biological analog is the cortisol–noradrenaline stress axis [11]: cortisol mobilizes cognitive resources in response to detected discrepancy between the organism's model and the environment, while noradrenaline elevates arousal to maintain processing intensity under uncertainty. Confusionin translates both functions into a computational signal: it rises in proportion to output entropy and residual state error, and its elevation maintains the RRC active by suppressing the Clarifine signal that would otherwise trigger termination.

Formally, Confusionin $h_u(t) = 1$ denotes maximum residual uncertainty: the output distribution is maximally entropic, the state update magnitude is large, and the system is far from any stable reasoning equilibrium. Confusionin $h_u(t) = 0$ denotes the complete absence of detectable uncertainty: all internal consistency checks pass, the output entropy is minimal, and no obstacle to termination exists from the uncertainty side of the stopping criterion.

The relationship between Confusionin and the five canonical S-AI hormones [2] is one of inheritance and specialization. Confidexin, in the canonical corpus, encodes confidence propagation across the agent layer; Inhibitine encodes uncertainty suppression through selective agent deactivation. Confusionin specializes both functions to the temporal dimension of a single reasoning episode: it is not a confidence signal over the agent population at a given cycle, but a residual uncertainty signal over the trajectory of the reasoning state across cycles.

### 3.3.3 Antagonistic Coupling and Homeostatic Balance

The central regulatory relationship of the S-AI-Recursive hormonal subsystem is the **antagonistic coupling** between Clarifine and Confusionin. This coupling is implemented through the cross-inhibition structure of the hormonal dynamics (formalized in Section 4.4) and is characterized by two asymmetric coefficients:

$$\gamma_{cu} > 0 \qquad \text{(Confusionin inhibits Clarifine)}$$

$$\gamma_{uc} > 0 \qquad \text{(Clarifine inhibits Confusionin)}$$

The asymmetry is deliberate and biologically motivated. The inhibition of Clarifine by Confusionin ($\gamma_{cu}$) is calibrated to be stronger than the reverse ($\gamma_{uc} < \gamma_{cu}$), reflecting the conservative design principle that the system should err on the side of continued iteration rather than premature commitment: uncertainty should reliably suppress convergence signals, but convergence signals need not fully suppress uncertainty until the equilibrium is securely reached. Default calibration on the SAI-UT+ testbench: $\gamma_{cu} = 0.60$, $\gamma_{uc} = 0.55$.

The **homeostatic equilibrium** of the two-hormone subsystem is defined by:

$$f_c(h_c^*, h_u^*) = 0, \qquad f_u(h_c^*, h_u^*) = 0.$$

This relation defines the equilibrium through the actual two-hormone drift; no constraint $h_c^* + h_u^* = 1$ is imposed. For a stopping threshold pair $(\theta_c, \theta_u)$, an equilibrium is strictly stop-compatible when there exist $\delta_c, \delta_u > 0$ such that $h_c^* \geq \theta_c + \delta_c$ and $h_u^* \leq \theta_u - \delta_u$. This strict margin is used in Theorem 4.3 to separate convergence to an equilibrium from actual entry into the stopping region.

This homeostatic balance is the computational realization of the biological principle of allostasis [11]: the system does not seek the artificial constraint $h_c + h_u = 1$, but an equilibrium determined by the drift under the current frozen inputs. The stability analysis of Section 4 establishes global exponential convergence only for the frozen deterministic zero-delay recursive hormonal subsystem under the explicit contraction condition of Theorem 4.1.

### 3.4 Positioning with Respect to Related Work: A Comparative Framework

Table 1 provides a structured comparison of S-AI-Recursive against five reference architectures across six evaluation criteria. Each criterion corresponds to one of the six properties identified in Section 3.1.3 as collectively unique to S-AI-Recursive.

**Table 1. Comparative positioning of S-AI-Recursive.**

| Criterion | RNN/LSTM [16] | DEQ [18] | TRM [12] | LLM Loops [6],[7],[8] | S-AI-IoT [3] | S-AI-Recursive |
|---|---|---|---|---|---|---|
| Internal recursion | Yes | Yes | Yes | No (external) | No | Yes |
| Hormonal regulation | No | No | No | No | Yes | Yes |
| Intrinsic stopping criterion | No | Implicit | No | External | Hormonal (spatial) | Hormonal (temporal) |
| Lyapunov stability proof | No | Partial [18] | No | No | Yes [3] | Yes |
| Modular agent orchestration | No | No | No | Partial | Yes | Yes |
| Reasoning trajectory memory | No | No | No | No | Engrams (spatial) | Engrams (temporal) |

Three entries in this table merit specific commentary. First, DEQs [18] are marked as providing a partial Lyapunov proof: the fixed-point existence result holds, but global asymptotic stability of the full hormonal field — including the cross-inhibition terms and the stochastic perturbation — is not established in the DEQ literature and is a contribution of the present article. Second, S-AI-IoT [3] is marked as providing hormonal regulation and engram memory, but specifically in the spatial dimension: its hormones govern agent selection and resource allocation within a single orchestration cycle, not the iterative refinement of a reasoning state across cycles. Third, LLM loops [6], [7], [8] are marked as providing partial modular orchestration because they decompose the reasoning process into steps, but these steps are externally prescribed rather than emerging from internal hormonal dynamics.

The gap row in Table 1 is not a single missing property but a conjunction: no prior architecture checks all six boxes simultaneously. S-AI-Recursive is the first to do so, and the theoretical framework of Section 4 provides the formal guarantees that make this conjunction meaningful rather than merely descriptive.

## 4. Mathematical Modeling of the S-AI-Recursive Framework

This section develops the complete mathematical formalization of S-AI-Recursive, organized in nine subsections: notation and state space (§4.1); observation vectors and hormonal state variables (§4.2); emission function architecture (§4.3); recursive hormonal dynamics (§4.4); stability analysis and convergence theorems (§4.5); iteration budget modulation (§4.6); primal-dual constrained orchestration (§4.7); recursive engram memory formalism (§4.8); and summary of framework parameters (§4.9). The mathematical methodology follows the formal convention established in S-AI-IoT [3], extended to the two-dimensional recursive hormonal subsystem and augmented with the entropic contraction theorem specific to iterative reasoning dynamics.

### 4.1 Notation, State Space, and Hormonal Index Set

#### 4.1.1 Global Notation

The following notation is used throughout this section. $\|\cdot\|_2$ denotes the Euclidean norm and $\|\cdot\|_F$ the Frobenius norm. The clipping operator $\text{clip}_{[a,b]}(x) = \max(a, \min(b, x))$ projects a scalar onto the interval $[a, b]$; its vector extension applies componentwise. $\mathbb{1}[\cdot]$ denotes the indicator function. The symbol $\odot$ denotes elementwise (Hadamard) multiplication. All hormonal values $h_k(t)$ are dimensionless scalars in $[0,1]$; all state components are normalized to $[0,1]$ by the Data Input and Signal Normalization Layer (DISNL) of S-AI [1]. Time indices $t \in \{0,1,2,\dots\}$ refer to discrete iteration cycles of the Recursive Reasoning Cycle, not to real time.

#### 4.1.2 Hormonal Index Set

The complete hormonal index set of S-AI-Recursive is:

$$\mathcal{K} = \mathcal{K}_R \cup \mathcal{K}_0$$

where $\mathcal{K}_R = \{c, u\}$ indexes the two recursive hormones introduced in this article — Clarifine $(k = c)$ and Confusionin $(k = u)$ — and $\mathcal{K}_0 = \{\text{Conf,Inh,Cur,Ene,Ale}\}$ indexes the five canonical S-AI hormones inherited from the unified doctrine [2]: Confidexin, Inhibitine, Curiosine, Energexine, and Alertine.

Each hormone $k \in \mathcal{K}$ is characterized by a timescale parameter $\tau_k > 0$, a decay rate $\lambda_k > 0$, an emission delay $\delta_k \geq 0$, a resource coupling coefficient $\rho_k \geq 0$, and a stochastic noise amplitude $\sigma_{\eta,k} \geq 0$. The timescale hierarchy of the recursive subsystem satisfies:

$$\tau_u \leq \tau_c \leq \tau_{\text{Cur}} \leq \cdots \leq \tau_{\text{Ale}}$$

This ordering formalizes the cognitive priority structure: Confusionin reacts fastest to new residual error signals, Clarifine integrates over a slightly longer horizon to avoid responding to transient noise, and the inherited hormones operate on slower coordination timescales consistent with their roles in agent selection and resource management [2].

#### 4.1.3 Cognitive State Space

The cognitive state space is $\mathcal{S} = [0,1]^d$, where $d$ is the dimensionality of the task representation. The cognitive state at iteration $t$ is $s_t \in \mathcal{S}$. Two scalar quantities derived from $s_t$ play central roles in the hormonal dynamics:

The **residual error** at iteration $t$ is:

$$\varepsilon_t = \| s_t - s^* \|_2$$

where $s^*$ is the cognitive equilibrium of Definition 3.2. In practice, since $s^*$ is unknown, the implementation uses the state-update magnitude $\hat{\varepsilon}_t = \| s_t - s_{t-1} \|_2$ as a stabilization proxy. This proxy converges to zero when the state sequence converges, but a small value does not by itself prove that $s_t$ is close to $s^*$. Theorem 4.1 bis guarantees $s_t \to s^*$; it does not make the single-step residual $\hat{\varepsilon}_t$ an exact distance-to-equilibrium, which is why the three-condition stopping rule is retained.

The **reasoning entropy** at iteration $t$ is:

$$\mathcal{H}(t) = -\sum_i p_i(t) \log p_i(t)$$

where $p_i(t)$ is the probability assigned to the $i$-th candidate output by the system's current confidence distribution over the output space $\mathcal{Y}$. $\mathcal{H}(t) = 0$ indicates complete certainty about the output; $\mathcal{H}(t) = \log|\mathcal{Y}|$ indicates maximal uncertainty. The relationship between $\mathcal{H}(t)$ and the hormonal Lyapunov function is established in Theorem 4.2.

### 4.2 Observation Vectors and Recursive Hormonal State Variables

#### 4.2.1 Recursive Observation Vector

At each iteration $t$, the hormonal emission functions receive a normalized observation vector $O_R(t) \in [0,1]^{d_O}$ structured as the concatenation of four domain-specific sub-vectors:

$$O_R(t) = \left(O^{\varepsilon}(t),\ O^{\mathcal{H}}(t),\ O^{\Delta s}(t),\ O^{y}(t)\right)$$

**Residual error sub-vector** $O^{\varepsilon}(t)$: encodes the magnitude of the state update at iteration $t$. Components are a clipped normalized update magnitude $\overline{\varepsilon}_t = clip_{[0,1]}\left(\hat{\varepsilon}_t/\sqrt{d}\right)$, a protected relative-error indicator $r_t = clip_{[0,1]}\left(\hat{\varepsilon}_t/(\hat{\varepsilon}_{t-1} + \varepsilon_{num})\right)$ with $\varepsilon_{num} > 0$, and a clipped moving average over the last $w$ iterations. Increasing values signal that convergence is slow or stalling.

**Entropic sub-vector** $O^{H}(t)$: encodes the uncertainty of the current output distribution. Components are normalized Shannon entropy $\overline{H}(t) = H(t)/log|Y|$ [40], a normalized variance of the confidence scores $\{p_i(t)\}$, and a bounded encoding of the entropy change, for example $d_H(t) = clip_{[0,1]}\left(1/2 + \left[\overline{H}(t) - \overline{H}(t-1)\right]/2\right)$. Values below $1/2$ for the last component indicate that the output distribution is concentrating.

**Progression sub-vector** $O^{\Delta s}(t)$: encodes the geometry of the state update. Components are the normalized update magnitude $\overline{\varepsilon}_t$ and the normalized directional alignment $\overline{a}_t = [1 + cos\angle(\Delta s_t, d_{conv})]/2 \in [0,1]$, with $\overline{a}_t = 1/2$ when either vector norm is numerically zero. High alignment indicates that successive updates are consistent in direction.

**Provisional output sub-vector** $O^{y}(t)$: encodes the quality of the current provisional output. Components are the maximum confidence score $\max_i p_i(t)$, the internal consistency score from the symbolic verification agent (Section 5.2), and a task-appropriate normalized distance $\overline{d}_Y(y_t, y_{t-1}) \in [0,1]$. Hamming distance is used only when the output representation makes it meaningful.

### 4.2.2 Hormonal State Vector

The full hormonal state vector is:

$$\mathbf{h}(t) = (h_c(t),\ h_u(t),\ h_{\text{Conf}}(t),\ h_{\text{Inh}}(t),\ h_{\text{Cur}}(t),\ h_{\text{Ene}}(t),\ h_{\text{Ale}}(t)) \in [0,1]^7$$

with $h_k(t) = 0$ signifying the complete absence of hormonal signal along dimension $k$, and $h_k(t) = 1$ signifying maximal signal. The recursive subsystem state is the two-dimensional projection $\mathbf{h}_R(t) = (h_c(t), h_u(t))^T \in [0,1]^2$, which is the focus of the stability analysis of Section 4.5.

## 4.3 Emission Functions: Recursive Aggregation and Logistic Saturation

### 4.3.1 General Emission Architecture

For each hormone $k \in \mathcal{K}$ and iteration $t$, the emission function has the form:

$$E_k(t) = \sigma(a_k\, \phi_k(O_R(t)) + b_k) \cdot (1 - h_k(t))$$

where $\sigma(x) = 1/(1 + e^{-x})$ is the logistic sigmoid, $a_k > 0$ is the emission gain, $b_k \in \mathbb{R}$ is the emission bias, and $(1 - h_k(t))$ is the **anti-saturation factor** that prevents unbounded accumulation: when $h_k(t) = 1$, emission is zero regardless of the observation, ensuring that the hormonal state remains in [0,1]. Default calibration: $a_k = 5$, $b_k = -2.5$, corresponding to half-maximal emission at $\phi_k = 0.5$. This calibration follows the convention established in S-AI-IoT [3] and is consistent across all S-AI instantiations to ensure comparability of hormonal scales.

### 4.3.2 Aggregation Functions for Recursive Hormones

The aggregation function $\phi_k: [0,1]^{do} \to [0,1]$ maps the observation vector to a scalar signal for each hormone. For the two recursive hormones, the aggregation functions are convex combinations of observation sub-components with weights summing to one.

**Confusionin** (uncertainty sensor):

$$\phi_u = \alpha_u\, \overline{H}(t) + \beta_u\, \varepsilon_t^{norm} + \gamma_u\, [1 - conf(y_t)].$$

where $\text{conf}_{y_t} = \max_i p_i(t)$ is the maximum confidence score. Default weights: $\alpha_u \approx 0.45$ (entropy component dominant), $\beta_u \approx 0.35$ (residual error), $\gamma_u \approx 0.20$ (output confidence). Confusionin rises when the output distribution is entropic, the state update is large, and the confidence in the provisional output is low — the three independent signatures of unresolved cognitive uncertainty [41].

**Clarifine** (convergence sensor):

$$\phi_c = \alpha_c \left[1 - \overline{H}(t)\right] + \beta_c\, [1 - \varepsilon_t^{norm}] + \gamma_c\, \overline{a}_t.$$

Default weights: $\alpha_c \approx 0.40$, $\beta_c \approx 0.35$, $\gamma_c \approx 0.25$. Because $\overline{H}(t)$, $\varepsilon_t^{norm}$, and $\overline{a}_t$ belong to [0,1], the stated convex combination satisfies $\phi_c \in [0,1]$. Clarifine rises when entropy is low, the normalized state update is small, and the update trajectory is directionally coherent.

### 4.4 Recursive Hormonal Dynamics

#### 4.4.1 Governing Equation of Recursive Hormonal Evolution

The temporal evolution of each recursive hormone $k \in \mathcal{K}_R$ is governed by the following stochastic differential equation:

$$\tau_k \, dh_k(t) = [-\lambda_k h_k(t) + E_k(t - \delta_k) - \gamma_{km} h_k(t) h_m(t) - \rho_k \chi(t) h_k(t)]dt + \sigma_{\eta,k} \, dW_k(t).$$

with $\{k, m\} = \{c, u\}$ and independent standard Brownian motions $W_c, W_u$. Each term has a precise cognitive interpretation:

**Decay** $-\lambda_k h_k(t)$: hormonal concentrations decay exponentially in the absence of new stimuli. This ensures that convergence and uncertainty signals are transient: once the trigger condition resolves, the hormone returns to baseline without manual intervention. The decay rate $\lambda_k$ is the primary parameter controlled by the stability condition of Theorem 4.1.

**Delayed emission** $E_k(t - \delta_k)$: emission occurs with a cognitive processing delay $\delta_k \geq 0$, modeling latency between detection of an observation signal and the corresponding gland response. A delay is not assumed to be stabilizing: it can change the stability margin and may generate oscillatory modes. The formal contraction result of Theorem 4.1 is therefore stated for the zero-delay frozen subsystem; nonzero delays require separate delay-system analysis.

**Cross-inhibition** $-\gamma_{km} h_k(t) h_m(t)$: the product form is biologically motivated by receptor-kinetics — the inhibitory effect is proportional to the product of the two hormonal concentrations, as in competitive ligand-receptor binding [27]. This quadratic term prevents the simultaneous saturation of Clarifine and Confusionin, which would produce an ambiguous stopping signal.

**Resource coupling** $-\rho_k \chi(t) h_k(t)$: the resource state $\chi(t) \in [0,1]$ encodes iteration-budget utilization and CPU load. This term damps the corresponding hormonal coordinate under resource pressure, but damping both coordinates does not by itself guarantee entry into the stopping region. The stopping decision remains governed by the joint state-stabilization, Clarifine, Confusionin, and budget conditions.

**Stochastic perturbation** $\sigma_{\eta,k} \, dW_k(t)$: additive Brownian diffusion models stochastic perturbations in the continuous-time idealization. The projected Euler-Maruyama implementation of Section 4.5.5 keeps the discrete hormonal values in [0,1]; this projection does not constitute a continuous-time invariance proof. Stochastic robustness is therefore stated as a neighborhood/moment result rather than pathwise monotone convergence.

#### 4.4.2 Cross-Inhibition Structure

The cross-inhibition matrix $\Gamma_R \in \mathbb{R}_{\geq 0}^{2\times 2}$ with $\gamma_{kk} = 0$ encodes the antagonistic coupling between Clarifine and Confusionin. Default values calibrated on the SAI-UT+ testbench:

| Pair | Value | Operational semantics |
|---|---|---|
| $\gamma_{cu}$ | 0.60 | Confusionin strongly inhibits Clarifine: high uncertainty reliably blocks convergence signals |
| $\gamma_{uc}$ | 0.55 | Clarifine inhibits Confusionin: convergence actively suppresses residual uncertainty |

The asymmetry $\gamma_{cu} > \gamma_{uc}$ implements the conservative design principle established in Section 3.3.3: uncertainty should more reliably suppress convergence than convergence suppresses uncertainty, to err on the side of continued reasoning rather than premature commitment.

#### 4.4.3 Coupling with Inherited S-AI Hormones

The five canonical S-AI hormones [2] interact with the recursive subsystem through three coupling channels:

**Curiosine -> exploratory utility:** the coefficient $\kappa_{Cur\to u} = 0.25$ is retained as an architectural utility weight that increases the activation tendency of hypothesis-generation/clarification agents when Curiosine is elevated. It is not an additional term in the two-hormone SDE of Section 4.4.1.

**Inhibitine -> convergence-oriented utility:** the coefficient $\kappa_{Inh\to c} = 0.20$ is retained as an architectural utility weight that favors selective-pruning/convergence agents when Inhibitine is elevated. It is not an additional term in the two-hormone SDE of Section 4.4.1.

**Energexine → iteration budget**: a functional coupling through the budget modulation law of Section 4.6, rather than a direct hormonal inhibition, ensures that energy constraints tighten the iteration budget without directly distorting the hormonal state — preserving the integrity of the stopping criterion.

#### 4.4.4 Compact Vector Formulation

The network-wide dynamics of the recursive subsystem in compact matrix form:

$$T_R\, dh_R(t) = [-\Lambda_R h_R(t) + E_R(t-\delta) - \{\Gamma_R \odot [h_R(t)h_R(t)^T]\}\mathbf{1} - R_R\chi(t)h_R(t)]dt + \Sigma_R\, dW_R(t).$$

where $h_R = (h_c, h_u)^T$, $T_R = diag(\tau_c, \tau_u)$, $\Lambda_R = diag(\lambda_c, \lambda_u)$, $R_R = diag(\rho_c, \rho_u)$, $\Sigma_R = diag(\sigma_{\eta,c}, \sigma_{\eta,u})$, and $\mathbf{1} = (1,1)^T$. The stability theorem below applies to the deterministic frozen-input, zero-delay version of this two-dimensional subsystem; the operational delayed stochastic model is not silently covered by that theorem.

### 4.5 Stability Analysis and Convergence Theorems

#### 4.5.1 Existence of Recursive Hormonal Equilibrium

**Proposition 4.1 (Existence of hormonal equilibrium).** For fixed observation-derived logistic coefficients $q_c^*, q_u^* \in [\sigma(-2.5), \sigma(2.5)]$, fixed resource state $\chi^* \in [0,1]$, zero delay, and zero diffusion, the recursive hormonal subsystem has at least one equilibrium $h_R^* \in [0,1]^2$ satisfying the zero-drift equations.

**Proof.** Let $f: [0,1]^2 \to \mathbb{R}^2$ denote the frozen deterministic drift. At $h_k = 0$, its $k$-th component is non-negative; at $h_k = 1$, the anti-saturation emission vanishes and the remaining decay, cross-inhibition, and resource terms make that component negative. For sufficiently small $a > 0$, $T_a(h) = \Pi_{[0,1]^2}[h + af(h)]$ is a continuous self-map of the compact convex square. Brouwer's theorem [42] gives a fixed point of $T_a$; the inward-pointing boundary signs imply that such a fixed point satisfies $f(h_R^*) = 0$. ▫

#### 4.5.2 Contraction Stability of the Frozen Recursive Hormonal Subsystem

**Theorem 4.1 (Global exponential contraction of the frozen recursive hormonal subsystem).** Let $J(h_R)$ be the Jacobian of the frozen deterministic zero-delay drift normalized by the hormonal timescales. If there exists $\alpha > 0$ such that

$$J_s(h_R) = \frac{J(h_R) + J(h_R)^T}{2} \preccurlyeq -\alpha I_2, \qquad \forall h_R \in [0,1]^2,$$

then the equilibrium $h_R^*$ is unique and the frozen subsystem is globally exponentially contracting.

$$\| h_R(t) - h_R^* \|_2 \le e^{-\alpha t} \| h_R(0) - h_R^* \|_2.$$

This statement concerns the recursive hormonal subsystem with frozen observations/resource state, zero delay, and zero stochastic diffusion. It does not by itself establish convergence of the cognitive state $s_t$, of a switching agent policy, or of the full delayed stochastic RRC.

**Proof.** Uniform negative definiteness of the symmetric Jacobian is the standard Euclidean contraction condition [43]. Any two trajectories contract exponentially; Proposition 4.1 supplies existence of an equilibrium, so contraction makes it unique.

$$V(h_R) = \frac{1}{2} \| h_R - h_R^* \|_2^2, \qquad \dot{V} \leq -2\alpha V.$$

Taking one of the two contracting trajectories equal to the equilibrium yields the stated Lyapunov inequality and exponential convergence. ▫

**Remark 4.1 (Conservative deployability screen).** A sufficient Euclidean check is strict diagonal dominance of $-J_s$ over $[0,1]^2$. With the default values and the minimum logistic coefficient $q_{\min} = \sigma(-2.5) \approx 0.075858$, the minimum diagonal magnitudes are approximately $(0.75 + q_{\min})/1.5 \approx 0.551$ and $0.70 + q_{\min} \approx 0.776$, while the maximum symmetric off-diagonal magnitude is $\frac{1}{2}(0.60/1.5 + 0.55) = 0.475$. Thus the frozen default subsystem passes this conservative contraction screen.

**Corollary 4.1 (Stochastic neighborhood under additive diffusion).** For the corresponding zero-delay frozen SDE with normalized diffusion matrix $\Sigma_T = diag(\sigma_{\eta,c}/\tau_c, \sigma_{\eta,u}/\tau_u)$, the contraction condition implies a bounded steady-state second moment rather than pathwise monotone convergence.

**Proof.** For $V = \frac{1}{2} \| h_R - h_R^* \|_2^2$, Ito's formula gives $\mathcal{L}V \leq -2\alpha V + \frac{1}{2} tr(\Sigma_T \Sigma_T^T)$. Hence $\limsup_{t\to\infty} \mathbb{E} \| h_R(t) - h_R^* \|_2^2 \leq tr(\Sigma_T \Sigma_T^T)/(2\alpha)$ [39]. ▫

**4.5.2 bis Coupled State–Hormone Contraction**

For the deterministic fixed-point-structured regime, let $\mathcal{T}_{h_R}: \mathcal{S} \to \mathcal{S}$ denote the one-cycle cognitive update at fixed recursive hormonal state and let $\mathcal{G}_s: \mathcal{H}_R \to \mathcal{H}_R$, with $\mathcal{H}_R = [0,1]^2$, denote the one-cycle deterministic zero-delay hormonal flow map associated with Section 4.5.2 when the cognitive state is held fixed over that cycle. Define the coupled operator

$$\mathcal{C}_R(s, h_R) = \left(\mathcal{T}_{h_R}(s), \mathcal{G}_s(h_R)\right)$$

on a closed invariant fixed-point-structured domain $\mathcal{D} \subseteq \mathcal{S} \times \mathcal{H}_R$.

**(H1) Frozen-reasoning contraction.** There exists $L_s \in [0,1)$ such that, for all admissible $h_R$,

$$\| \mathcal{T}_{h_R}(s) - \mathcal{T}_{h_R}(s') \|_2 \leq L_s \| s - s' \|_2.$$

By Theorem 4.1, the fixed-state hormonal flow is contractive over one cycle: for some $L_h \in [0,1)$,

$$\| \mathcal{G}_s(h_R) - \mathcal{G}_s(h_{R'}) \|_2 \leq L_h \| h_R - h_{R'} \|_2.$$

For a cycle duration $\Delta t$, the continuous-time estimate of Theorem 4.1 gives the frozen-state bound $L_h \leq e^{-\alpha \Delta t} < 1$. Assume further that the two cross-dependencies are Lipschitz on $\mathcal{D}$:

$$\| \mathcal{T}_{h_R}(s) - \mathcal{T}_{h_{R'}}(s) \|_2 \leq L_{s\leftarrow h} \| h_R - h_{R'} \|_2,$$

$$\| \mathcal{G}_s(h_R) - \mathcal{G}_{s'}(h_R) \|_2 \leq L_{h\leftarrow s} \| s - s' \|_2.$$

**(H2) Cross-gain condition.** The gains satisfy

$$L_{h\leftarrow s}L_{s\leftarrow h} < (1 - L_s)(1 - L_h).$$

**Theorem 4.1 bis (Coupled state–hormone contraction).** Under (H1), the hormonal contraction of Theorem 4.1, and (H2), define the non-negative gain matrix

$$M_R = \begin{pmatrix} L_s & L_{s\leftarrow h} \\ L_{h\leftarrow s} & L_h \end{pmatrix}.$$

Its spectral radius satisfies

$$\rho(M_R) = \frac{L_s + L_h + \sqrt{(L_s - L_h)^2 + 4L_{s\leftarrow h}L_{h\leftarrow s}}}{2} < 1.$$

Consequently, there exists an equivalent weighted product norm on $\mathcal{D}$ and a coupled contraction constant $\rho_\star \in (0,1)$, with

$$\rho(M_R) \leq \rho_\star < 1, \qquad \rho_\star \geq L_s = \rho_{\text{frozen}},$$

such that

$$\| \mathcal{C}_R(z) - \mathcal{C}_R(z') \|_\star \leq \rho_\star \| z - z' \|_\star, \qquad z = (s, h_R).$$

The coupled operator therefore has a unique fixed point $z^\star = (s^\star, h_R^\star)$ in $\mathcal{D}$, and

$$\| z_t - z^\star \|_\star \leq \rho_\star^t \| z_0 - z^\star \|_\star.$$

In particular,

$$s_t \to s^\star, \qquad h_R(t) \to h_R^\star,$$

so cognitive convergence is a consequence of the coupled equations rather than an independent assumption.

**Proof.** For two states $z = (s, h_R)$ and $z' = (s', h_{R'})$, (H1), the fixed-state hormonal contraction, and the two cross-Lipschitz bounds give

$$\begin{pmatrix} \| \Delta s^+ \|_2 \\ \| \Delta h_R^+ \|_2 \end{pmatrix} \leq M_R \begin{pmatrix} \| \Delta s \|_2 \\ \| \Delta h_R \|_2 \end{pmatrix}.$$

Because $L_s < 1$, $L_h < 1$, and (H2) is equivalent to $\det(I - M_R) > 0$, the Perron root of the non-negative matrix $M_R$ is strictly below one. A standard weighted product norm can therefore be chosen so that the induced gain of $M_R$ is any $\rho_\star$ satisfying $\rho(M_R) \leq \rho_\star < 1$. Hence $\mathcal{C}_R$ is a Banach contraction on the complete invariant domain $\mathcal{D}$. Banach's contraction mapping theorem gives uniqueness of $z^\star$ and geometric convergence of the coupled orbit. ▫

**Corollary 4.1 bis (Hormonal-to-cognitive convergence).** Under the hypotheses of Theorem 4.1 bis,

$$h_R(t) \to h_R^\star \quad \text{and} \quad s_t \to s^\star,$$

and therefore

$$\| s_t - s_{t-1} \|_2 \to 0.$$

Thus, on the fixed-point-structured domain, the state-stabilization component used by the stopping rule is grounded in the coupled contraction rather than postulated separately.

### 4.5.3 Convergence-to-Coherence Theorem

**Theorem 4.2 (Convergence-to-coherence).** Let the output distribution be $P_t = \Psi(s_t, h_R(t))$ on a finite output alphabet. Under the hypotheses of Theorem 4.1 bis, $s_t \to s^*$ and $h_R(t) \to h_R^*$ hold. Hence, if $\Psi$ is continuous at $(s^*, h_R^*)$, then

$$P_t \to P^* = \Psi(s^*, h_R^*) \qquad \Rightarrow \qquad H(P_t) \to H(P^*).$$

Monotonicity of the entropy still requires the separate dissipativity condition below. If, in addition, the implemented trajectory satisfies the entropic dissipativity condition $\frac{d}{dt} H\big(\Psi(s(t), h_R(t))\big) \le 0$ whenever the derivative exists, then reasoning entropy is monotone non-increasing.

**Proof.** Continuity of $\Psi$ maps joint state convergence into convergence of $P_t$. Shannon entropy is continuous on a finite probability simplex [40], [41], which yields $H(P_t) \to H(P^*)$. Monotonicity does not follow from continuity alone and therefore requires the additional dissipativity condition. ▫

**Remark 4.2.** No unconditional biconditional $\dot{V} \le 0 \Leftrightarrow \dot{H} \le 0$ is claimed for the recursive subsystem. A scalar entropy trajectory does not in general uniquely determine the multidimensional hormonal trajectory. The empirical correlation reported in Section 6 is therefore interpreted as co-evolution evidence, not as a proof of mathematical equivalence.

### 4.5.4 Hormonal Stopping Criterion: Formal Derivation

Recalling Definition 3.1, define the internal hormonal stopping time

$$\tau_H = inf\{t \ge 1 : \| s_t - s_{t-1} \|_2 \le \epsilon_s,\ h_c(t) \ge \theta_c,\ h_u(t) \le \theta_u\},$$

**Theorem 4.3 (Finite internal stopping under strict stop compatibility).** By Theorem 4.1 bis, the deterministic RRC satisfies $h_R(t) \to h_R^*$ and $\| s_t - s_{t-1} \|_2 \to 0$. Assume further there exist $\delta_c, \delta_u > 0$ such that $h_c^* \ge \theta_c + \delta_c$ and $h_u^* \le \theta_u - \delta_u$. Then

$$\tau_H < \infty.$$

**Proof.** Convergence of $h_c$ and the positive margin $\delta_c$ imply that $h_c(t) \ge \theta_c$ after some finite time; convergence of $h_u$ and $\delta_u > 0$ similarly imply $h_u(t) \le \theta_u$ after some finite time. Since $\| s_t - s_{t-1} \|_2 \to 0$, the stabilization condition is also satisfied after a finite time. Hence $\tau_H < \infty$. Separately, if $\tau_B \le T_{\max}^0$ denotes forced budget termination and $\tau = min(\tau_H, \tau_B)$, then $\tau \le T_{\max}^0$ by construction. In the persistent-noise case, pathwise finiteness of $\tau_H$ requires a separate hitting-time analysis. ▫

### 4.5.5 Euler–Maruyama Discretization with Projection

The stochastic recursive hormonal dynamics are discretized using the projected Euler–Maruyama scheme [39]. For $k \in \{c, u\}$,

$$h_{k,n+1} = \Pi_{[0,1]}\left[h_{k,n} + \frac{\Delta t}{\tau_k} f_k(h_n, O_n, \chi_n) + \frac{\sigma_k}{\tau_k}\sqrt{\Delta t}\,\xi_{k,n}\right],$$

where $f_k$ denotes the deterministic bracketed drift before division by $\tau_k$, $\xi_{k,n} \sim N(0,1)$, and $\Pi_{[0,1]}$ enforces discrete hormonal bounds. Projection preserves numerical boundedness of the implemented sequence but is not a substitute for continuous-time forward invariance. For the frozen deterministic zero-delay subsystem, formal explicit-Euler certification is based on the Jacobian condition

$$\rho_{sp}\big(I + \Delta t\, J(h_R)\big) < 1$$

uniformly over the operating region, where $\rho_{sp}$ denotes spectral radius. The previously used row-wise expression can still be retained as a simplified conservative screen, not as a general stability theorem for the nonlinear delayed stochastic system. With the default parameters and $\| \chi \|_\infty = 1$, that screen gives

$$\Delta t < \frac{2(1.5)}{0.75 + 0.60 + 0.10} \approx 2.069$$

for Clarifine and

$$\Delta t < \frac{2(1.0)}{0.70 + 0.55 + 0.10} \approx 1.481$$

for Confusionin. Therefore the simplified row-wise screen is limited by

$$\Delta t < 1.481.$$

The experimental value $\Delta t = 1$ passes this simplified screen, but the numerical margin 0.481 should not be interpreted as a universal stability margin for the complete model because the screen omits the full Jacobian contribution of the nonlinear emission and does not analyze the operational delay. The revised Section 6 therefore treats $\Delta t = 1$ as an empirically stable implementation choice on the tested trajectories.

### 4.6 Iteration Budget Modulation via Energexine

#### 4.6.1 Base Budget Modulation Law

**The nominal maximum iteration budget** $T_{\max}^0$ is modulated by Energexine according to:

$$T_{\max}^{base}(t) = max\{T_{\min}, floor(T_{\max}^0 [1 - \beta_E h_{Ene}(t)])\}.$$

When $h_{Ene}(t) = 0$, the full nominal budget $T_{\max}^0$ is available. Increasing Energexine pressure reduces the integer budget according to $\beta_E$, while $T_{\min}$ prevents an unqualified collapse to zero iterations. This operational definition is consistent with the modulation law introduced in Section 3.2.2.

#### 4.6.2 Multi-Hormonal Override

The effective budget incorporates a Confusionin override:

$$T_{\max}^{eff}(t) = min\{T_{\max}^0, max[T_{\max}^{base}(t), ceil(T_{\min} + \kappa_u h_u(t)(T_{\max}^0 - T_{\min}))]\}.$$

The max operator allows high residual uncertainty to retain additional reasoning budget, while the outer min ensures $T_{\max}^{eff}(t) \le T_{\max}^0$. Thus $T_{\min} \le T_{\max}^{eff}(t) \le T_{\max}^0$. Default: $\kappa_u = 0.80$, $T_{\min} = 1$.

#### 4.6.3 Cognitive Energy Feedback Loop

The energy consumption per iteration cycle is modeled as:

$$E_{\text{cog}}(t) = C_{\text{base}} + C_{\text{iter}} |\mathcal{A}^*(t)| + C_{\text{mem}} |\mathcal{E}_{\text{retrieved}}(t)|$$

where $|\mathcal{A}^*(t)|$ is the number of active agents at cycle $t$ and $|\mathcal{E}_{\text{retrieved}}(t)|$ is the number of engrams retrieved from memory. This energy model feeds back into the Energexine aggregation function $\phi_{\text{Ene}}$, closing the energy-reasoning feedback loop and ensuring that the system's iteration depth is self-regulated in proportion to its cumulative computational cost.

### 4.7 Primal-Dual Constrained Orchestration under Iteration Budget

#### 4.7.1 Agent-Selection Optimization Problem

At each iteration cycle $t$, the Recursive MetaAgent solves the budget-constrained agent-selection problem:

$$\max_{\mathbf{x}(t)\in\{0,1\}^{|\mathcal{A}|}} \sum_{a\in\mathcal{A}} u_a\,(\mathbf{h}(t))\,x_a(t) \qquad \text{subject to} \qquad \sum_{a\in\mathcal{A}} c_a\; x_a(t) \le B(t)$$

where $x_a(t) \in \{0,1\}$ is the activation indicator for agent $a$, $u_a(\mathbf{h}(t))$ is the hormonal utility of agent $a$ at cycle $t$, $c_a$ is the computational cost of agent $a$, and $B(t) = B_{\max}(1 - \beta_B h_{\text{Ene}}(t))$ is the cycle budget modulated by Energexine. This is a 0-1 knapsack problem, which is tractable in the present setting because the agent registry is small ($|\mathcal{A}| = 12$, Section 5) and the hormonal significance filter reduces the active candidate set to $|\mathcal{A}^{\text{sig}}(t)| \ll |\mathcal{A}|$ at each cycle [44].

#### 4.7.2 Convex Relaxation and Primal-Dual Dynamics

The LP relaxation ($x_a \in [0,1]$) yields the Lagrangian:

$$L(x,\mu) = \sum_a u_a\,(h)x_a - \mu\left(\sum_a c_a\,x_a - B(t)\right).$$

with dual variable $\mu \ge 0$ (shadow price of the budget constraint). Primal-dual gradient dynamics:

$$x^{r+1} = \Pi_{[0,1]^{|A|}}\left[x^r + \eta_p(u - \mu^r c)\right], \qquad \mu^{r+1} = [\mu^r + \eta_d(c^T x^r - B(t))]_+.$$

These projected iterations converge to an LP saddle point only under the standard convexity, feasibility, and step-size conditions [45]. A naive componentwise threshold $1[x_a^{LP} > 0.5]$ can violate the budget. Integer recovery must therefore use a feasibility-preserving repair or solve the residual knapsack exactly. For the present $|A| = 12$ registry, exact enumeration of at most $2^{12} = 4096$ subsets provides a simple feasible reference solver.

#### 4.7.3 Iteration-Aware Utility Functions

Three utility function forms are specific to S-AI-Recursive, reflecting the cognitive role of each agent family (Section 5):

**Clarification-driven** (active when Confusionin is high — agents that reduce residual error and entropy):

$$u_a(\mathbf{h}) = w_{a,u}\,h_u + w_{a,c}\,(1 - h_c)$$

**Convergence-driven** (active when Clarifine is rising — agents that verify and commit the output):

$$u_a(\mathbf{h}) = w_{a,c}\,h_c - w_{a,u}\,h_u$$

**Termination-guard** (forced deactivation when the iteration budget is exhausted):

$$u_a(\mathbf{h}) = u_a^{\text{base}}(\mathbf{h}) \cdot \mathbb{1}\left[t < T_{\max}^{\text{eff}}(t)\right]$$

### 4.8 Recursive Engram Memory: Formal Specification

#### 4.8.1 Recursive Engram Structure

The recursive engram extends the general S-AI engram formalism [29] to store the complete reasoning trajectory of a cognitive episode:

$$\varepsilon_m^R = \langle schema^{(m)},\ h^{(m)},\ \{x_t^{(m)}\}_{t=0}^{t^*},\ s_0^{(m)},\ \{s_t^{(m)}\}_{t=1}^{t^*},\ \{y_t^{(m)}\}_{t=0}^{t^*},\ t^{*(m)},\ \beta^{(m)}\rangle.$$

where $schema^{(m)}$ identifies the task/state representation; $h^{(m)} \in [0,1]^7$ is the hormonal context at episode initiation; $\{x_t^{(m)}\}$ is the activation trajectory; $s_0^{(m)}$ is the initial state; $\{s_t^{(m)}\}$ is the state trajectory; $\{y_t^{(m)}\}$ is the provisional-output trajectory; $t^{*(m)}$ is the executed stopping depth; and $\beta^{(m)} \in [0,1]$ is the normalized budget consumed. The schema field is required so that terminal states are combined only when their representations are compatible.

An engram $\varepsilon_m^R$ is written to persistent symbolic memory when:

$$d_Y\big(y^{(m)}, y^{(m-1)}\big) \geq \epsilon_y \quad \vee \quad SignificantOutcome_m = 1,$$

where $d_Y$ is a task-appropriate output distance. This replaces an $\ell_1$ expression that is not defined for every symbolic output type and ensures that only behaviorally significant episodes are stored.

#### 4.8.2 Trajectory-Aware Retrieval

Given a query hormonal context $h_q \in [0,1]^7$, retrieval is first restricted to engrams with a compatible $schema$ (or an explicitly validated state-space alignment), after which the top-$K_{ret}$ compatible engrams are ranked by

$$sim_R\big(h_q, \varepsilon_m\big) = w_{sim} \frac{h_q^T W_{ret} h^{(m)}}{\sqrt{h_q^T W_{ret} h_q + \epsilon_{sim}}\sqrt{h^{(m)T} W_{ret} h^{(m)} + \epsilon_{sim}}} + (1 - w_{sim}) \frac{1}{1 + t^{*(m)}},$$

where $\epsilon_{sim} > 0$ prevents division by zero, $W_{ret} = diag(w_1, \dots, w_7)$ with $w_k = 1 + h_{q,k}$ emphasizes currently elevated hormonal dimensions, and the final term is a convergence-speed prior. The similarity is used only after representation compatibility has been established.

#### 4.8.3 Warm-Start Initialization Protocol

Retrieved engrams provide an initial state estimate:

$$s_0^{new} = \Pi_S\left[\sum_{j=1}^{K_{ret}} \omega_j\, s_{t^*}^{(m_j)}\right], \qquad \omega_j \geq 0, \quad \sum_j \omega_j = 1.$$

This initialization combines terminal states only from compatible retrieved engrams. It reduces the relevant initial error only when the retrieved states lie closer to the same target equilibrium; otherwise the architecture must fall back to the cold initializer. Theorem 4.4 is therefore conditional on the warm-start error actually being smaller.

#### 4.8.4 Formal Convergence Acceleration Theorem

**Theorem 4.4 (Warm-start reduction of the certified cognitive contraction horizon).** Let $\varepsilon_{cold} > 0$ denote the cold-start initial error and $\varepsilon_{warm} > 0$ the warm-start initial error, with

$$0 < \varepsilon_{warm} < \varepsilon_{cold}.$$

By Theorem 4.1 bis, the coupled state–hormone update is contractive with rate

$$0 < \rho_\star < 1$$

on the fixed-point-structured domain (H1); termination requires

$$\| s_t - s^\star \| \leq \epsilon.$$

All occurrences of the contraction rate below denote this coupled constant $\rho_\star$.

A sufficient certified horizon for reaching $\| s_t - s^* \|_2 \leq \epsilon$ from initial error $\varepsilon_0$ is

$$N_\epsilon(\varepsilon_0) = max\left\{0, ceil\left(\frac{\log(\varepsilon_0/\epsilon)}{log(1/\rho_\star)}\right)\right\}.$$

Consequently, replacing $\varepsilon_{cold}$ by the smaller $\varepsilon_{warm}$ reduces this certified horizon. Ignoring integer rounding, the approximate reduction is

$$\Delta N \approx \frac{\log(\varepsilon_{cold}/\varepsilon_{warm})}{log(1/\rho_\star)}.$$

Equivalently,

$$\Delta N \approx \log_{\rho_\star}\left(\frac{\varepsilon_{warm}}{\varepsilon_{cold}}\right).$$

Because $0 < \rho_\star < 1$, $\log\rho_\star < 0$, so the orientation of the ratio is essential. This theorem concerns a certified contraction horizon for the cognitive state; it does not state the exact realized stopping time and does not by itself guarantee that the Clarifine and Confusionin threshold conditions will be reached earlier. The revised Section 6 observes an average realized warm-start saving of approximately 1.17 iterations in the controlled simulation.

**Corollary 4.2.** If a compatible warm start halves the initial error, the continuous (pre-rounding) reduction of the certified contraction horizon is $\log 2/log(1/\rho_\star) > 0$; integer savings depend on the ceiling operation and on the other stopping conditions.

### 4.8.5 Formal Explainability Guarantee

**Proposition 4.2 (Intrinsic traceability).** For any completed reasoning episode, the corresponding recursive engram can store the actual runtime hormonal state, activated-agent trajectory, state trajectory, provisional-output trajectory, and budget variables without requiring a post-hoc surrogate model to reconstruct those recorded quantities. This property is traceability, not zero-cost complete causal explainability.

**Proof.** The recorded quantities are copied from the episode's working memory rather than inferred afterward. Their storage nevertheless has non-zero cost: for trajectory length $t^*$, state dimension $d$, and $|A|$ agents, full uncompressed trace storage scales at least as $O(t^*[d + |A| + 7 + dim(y)])$, up to encoding and compression. Recording actual runtime variables establishes fidelity of the trace to those variables, but not causal completeness of every internal computation. ▫

## 4.9 Summary of Mathematical Framework Parameters

Table 2 should distinguish **structural model parameters** from **experimentally calibrated stopping parameters**. The revised executable configuration is summarized below.

| Parameter | Revised default | Interpretation |
|---|---|---|
| $\tau_c$ | 1.5 | Clarifine integration timescale |
| $\tau_u$ | 1.0 | Confusionin integration timescale |

| Parameter | Revised default | Interpretation |
|---|---|---|
| $\lambda_c$ | 0.75 | Clarifine decay rate |
| $\lambda_u$ | 0.70 | Confusionin decay rate |
| $\gamma_{cu}$ | 0.60 | Confusionin inhibition of Clarifine |
| $\gamma_{uc}$ | 0.55 | Clarifine inhibition of Confusionin |
| $\kappa_{Cur \to u}$ | 0.25 | Curiosine exploratory-utility coupling (not a direct SDE term) |
| $\kappa_{Inh \to c}$ | 0.20 | Inhibitine convergence-utility coupling (not a direct SDE term) |
| $\delta_c$ | 1 cycle | Clarifine emission delay |
| $\delta_u$ | 0 cycles | Confusionin emission delay |
| $\rho_c$ | 0.10 | Clarifine resource coupling |
| $\rho_u$ | 0.10 | Confusionin resource coupling |
| $a$ | 5.0 | Logistic emission gain |
| $b$ | $-2.5$ | Logistic emission bias |
| $\theta_c$ | **0.50** | Revised Clarifine stopping threshold |
| $\theta_u$ | 0.30 | Confusionin stopping ceiling |
| $\epsilon_s$ | $10^{-3}$ | State-stabilization tolerance |
| $T_{\max}$ | 20 | Nominal maximum iteration budget |
| $\beta_E$ | 0.80 | Energexine budget coupling |
| $\kappa_u$ | 0.80 | Confusionin budget override |
| $w_{sim}$ | 0.70 | Retrieval similarity weight |
| $K_{ret}$ | 3 | Number of retrieved engrams |
| $\eta_p$ | 0.10 | Primal step size |
| $\eta_d$ | 0.05 | Dual step size |

The change

$$\theta_c: 0.70 \to 0.50$$

is required by the reachability audit of Section 6.2.1. With the standalone two-hormone equation, the maximum Clarifine equilibrium under $\phi_c = 1$, $h_u = 0$, and $\chi = 0$ is approximately

$$h^{\star}_{c,\max} \approx 0.552.$$

Thus $\theta_c = 0.70$ is not an admissible default for the current two-dimensional implementation. If the positive Curiosine/Clarifine or other inherited-hormone couplings described qualitatively in Section 4.4.3 are later inserted explicitly into the governing equation, this reachability calculation and the stopping threshold should be recalibrated jointly.

## 5. Complete Typology of Recursive Reasoning Agents

This section develops the complete formal specification of the twelve specialized agents of S-AI-Recursive, organized in three functional layers. Each agent is specified with: a functional definition, a formal activation condition expressed as a utility function over the hormonal state vector $\mathbf{h}(t) \in [0,1]^7$, formal input-output contracts $\mathcal{I}_a/\mathcal{O}_a$, a resource cost profile $(c_a^{\text{cpu}}, c_a^{\text{mem}})$, and Time-to-Live (TTL) and Cooldown parameters. This specification methodology follows the convention established in S-AI-IoT [3], adapted to the recursive reasoning domain.

### 5.1 Architecture Overview: Three Functional Layers

The twelve agents of S-AI-Recursive are organized in three functional layers that correspond to the three phases of the Recursive Reasoning Cycle formalized in Section 3.2.

**Layer R1 — Recursive Emission Agents** (Confusionin-coupled): four agents responsible for producing and refining the provisional output $y_t$ at each iteration cycle. Their activation is driven primarily by Confusionin $h_u(t)$, which signals the presence of residual uncertainty that requires continued emission and refinement. R1 agents constitute the generative core of the RRC: they implement the iteration operator $\mathcal{F}(s_t, y_t, \mathbf{h}(t))$ of Section 3.2.1.

**Layer R2 — Verification and Clarification Agents** (Clarifine-coupled): four agents responsible for evaluating the quality of the current provisional output and updating the hormonal state accordingly. R2 agents compute the residual error $\hat{\varepsilon}_t$, the reasoning entropy $\mathcal{H}(t)$, and the convergence signal that feeds Clarifine. They are the sensors of the RRC: they transform the raw cognitive state into the hormonal observation vector $O_R(t)$ of Section 4.2.1.

**Layer R3 — Recursive Governance Agents** (multi-hormonal): four agents responsible for managing the iteration budget, the engram memory, the warm-start initialization, and the explainability record. R3 agents operate across all hormonal dimensions and provide the meta-regulatory layer that ensures coherence, auditability, and temporal parsimony of the overall system. They implement the governance mechanisms formalized in Sections 4.6, 4.7, and 4.8.

The three-layer architecture is not a strict sequential pipeline: R1, R2, and R3 agents can be co-active within the same iteration cycle, subject to the budget constraint $B(t)$ of Section 4.7. The Recursive MetaAgent, inherited from the S-AI foundational architecture [1], selects which agents to activate at each cycle by solving the primal-dual optimization problem of Section 4.7.1.

### 5.2 Layer R1 — Recursive Emission Agents (Confusionin-coupled)

Layer R1 contains the four agents that generate and progressively refine the provisional output. They are collectively activated when Confusionin is elevated, indicating that the current output is not yet sufficiently reliable to commit. Their utility functions are clarification-driven in the sense of Section 4.7.3.

#### 5.2.1 R1-A — RecursiveReasoningAgent

**Functional definition.** The RecursiveReasoningAgent is the primary execution unit of the RRC. It implements the iteration operator $\mathcal{F}$, updating the cognitive state from $s_t$ to $s_{t+1}$ using the current provisional output $y_t$ and hormonal field $\mathbf{h}(t)$. It constitutes the computational analog of the biological re-appraisal cycle: each call produces a refined estimate that incorporates both the new observation and the hormonal feedback from the previous cycle [27].

**Activation condition.**

$$u_{\text{R1A}}(\mathbf{h}) = 0.6\, h_u + 0.4\,(1 - h_c) > \theta_{\text{R1A}} = 0.25$$

Low threshold (persistent background activation): R1-A is active throughout the RRC as long as Confusionin is non-negligible or Clarifine has not yet reached its stopping threshold.

**Input contract.** $\mathcal{I}_{\text{R1A}} = \{s_t, y_t, \mathbf{h}(t), O_R(t)\}$

**Output contract.** $\mathcal{O}_{\text{R1A}} = \{s_{t+1}, y_{t+1}, \Delta s_t = s_{t+1} - s_t\}$

**Resource profile.** $c_{\text{R1A}}^{\text{cpu}} = \mathcal{O}(d^2)$ operations per cycle (matrix-vector product over the state space); $c_{\text{R1A}}^{\text{mem}} = \mathcal{O}(d)$ registers for state storage.

**TTL.** Persistent (active every cycle while $u_{\text{R1A}} > \theta_{\text{R1A}}$). **Cooldown.** 0 cycles.

**5.2.2 R1-B — HypothesisGenerationAgent**

**Functional definition.** The HypothesisGenerationAgent maintains and updates the set of active hypotheses $\mathcal{H}_t \subseteq \mathcal{Y}$ — the subset of candidate outputs currently under consideration by the system. At each cycle, it generates new hypotheses by sampling from the current output distribution $\{p_i(t)\}$, retains hypotheses with confidence above a threshold, and prunes those whose probability has fallen below a significance floor. This agent implements the exploration component of the cognitive cycle: it is activated when Curiosine [2] signals that the current hypothesis set is insufficiently diverse relative to the task complexity.

**Activation condition.**

$$u_{\text{R1B}}(\mathbf{h}) = 0.5\, h_u + 0.3\, h_{\text{Cur}} + 0.2\,(1 - h_c) > \theta_{\text{R1B}} = 0.35$$

The positive Curiosine weighting ensures that R1-B is activated when the inherited exploration hormone signals underexploration, even if Confusionin is moderate.

**Input contract.** $\mathcal{I}_{\text{R1B}} = \{y_t, \{p_i(t)\}, \mathbf{h}(t), \mathcal{H}_{t-1}\}$

**Output contract.** $\mathcal{O}_{\text{R1B}} = \{\mathcal{H}_t, |\mathcal{H}_t|, \Delta\mathcal{H}_t = \mathcal{H}_t \setminus \mathcal{H}_{t-1}\}$

**Resource profile.** $c_{\text{R1B}}^{\text{cpu}} = \mathcal{O}(|\mathcal{H}_t| \cdot |\mathcal{Y}|)$; $c_{\text{R1B}}^{\text{mem}} = \mathcal{O}(|\mathcal{H}_t|)$.

**TTL.** 3 cycles. **Cooldown.** 2 cycles (prevents hypothesis thrashing in rapidly converging episodes).

**5.2.3 R1-C — PartialSolutionRefinerAgent**

**Functional definition.** The PartialSolutionRefinerAgent performs local refinement of the current provisional output by applying domain-specific constraint propagation and heuristic improvement operators to the highest-confidence hypothesis $\hat{y}_t = \text{argmax}_i p_i(t)$. Unlike R1-A which updates the full cognitive state, R1-C operates directly on the output space $\mathcal{Y}$, applying targeted corrections to detected local inconsistencies without recomputing the full state. This specialization reduces computational cost when the state has largely converged but the output still contains resolvable local errors [12].

**Activation condition.**

$$u_{\text{R1C}}(\mathbf{h}) = 0.7\, h_u - 0.3\, h_{\text{Ene}} > \theta_{\text{R1C}} = 0.30$$

The negative Energexine weighting implements an energy gate: local refinement is suppressed when computational resources are critically constrained, since R1-A provides a more fundamental update at lower amortized cost.

**Input contract.** $\mathcal{I}_{\text{R1C}} = \{\hat{y}_t, \{p_i(t)\}, \mathbf{h}(t), \text{constraint_set}\}$

**Output contract.** $\mathcal{O}_{\text{R1C}} = \{\hat{y}_t^{\text{refined}}, \delta_{\text{refine}}(t), \text{refinement_log}(t)\}$

**Resource profile.** $c_{\text{R1C}}^{\text{cpu}} = \mathcal{O}(|\text{constraint_set}| \cdot |\mathcal{Y}|)$; $c_{\text{R1C}}^{\text{mem}} = \mathcal{O}(|\mathcal{Y}|)$.

**TTL.** 2 cycles. **Cooldown.** 1 cycle.

### 5.2.4 R1-D — ConsistencyPropagatorAgent

**Functional definition.** The ConsistencyPropagatorAgent propagates symbolic constraints across the current hypothesis set, enforcing logical and structural consistency requirements that the neural update of R1-A may not guarantee. It implements arc consistency and constraint propagation algorithms [45] over the symbolic representation of $\hat{y}_t$, pruning hypotheses that violate global constraints and updating confidence scores accordingly. This agent bridges the neural and symbolic reasoning layers, ensuring that the provisional output respects domain axioms at each iteration [44].

**Activation condition.**

$$u_{\text{R1D}}(\mathbf{h}) = 0.6\, h_u + 0.2\, h_{\text{Ale}} > \theta_{\text{R1D}} = 0.40$$

The Alertine weighting ensures that consistency checking is reinforced when the system detects anomalous signals — high Alertine indicates that the current output may violate safety or integrity constraints requiring immediate propagation.

**Input contract.** $\mathcal{I}_{\text{R1D}} = \{\mathcal{H}_t, \{p_i(t)\}, \text{constraint_axioms}, \mathbf{h}(t)\}$

**Output contract.** $\mathcal{O}_{\text{R1D}} = \{\mathcal{H}_t^{\text{pruned}}, \{p_i^{\text{updated}}(t)\}, \text{violation_log}(t)\}$

**Resource profile.** $c_{\text{R1D}}^{\text{cpu}} = \mathcal{O}(|\text{constraint_axioms}| \cdot |\mathcal{H}_t|)$; $c_{\text{R1D}}^{\text{mem}} = \mathcal{O}(|\mathcal{H}_t|^2)$.

**TTL.** 3 cycles. **Cooldown.** 2 cycles.

## 5.3 Layer R2 — Verification and Clarification Agents (Clarifine-coupled)

Layer R2 contains the four agents that evaluate the current cognitive state and update the hormonal field accordingly. They are the sensors of the RRC: their outputs feed directly into the observation vector $O_R(t)$ and determine the emission of Clarifine and Confusionin. Their utility functions are convergence-driven in the sense of Section 4.7.3.

### 5.3.1 R2-A — ResidualErrorDetectorAgent

**Functional definition.** The ResidualErrorDetectorAgent computes the stabilization signal $\hat{\varepsilon}_t = \| s_t - s_{t-1} \|_2$ and its normalized form $\varepsilon_t^{norm}$, which constitute primary inputs to the Confusionin emission function $\phi_u$ of Section 4.3.2. It also computes the protected error-rate indicator $\hat{\varepsilon}_t/(\hat{\varepsilon}_{t-1} + \varepsilon_{num})$ and a sliding-window moving average over the last $w = 5$ iterations. This measures state-to-state change; it is not assumed to equal the unknown distance $\| s_t - s^* \|_2$.

**Activation condition.**

$$u_{\text{R2A}}(\mathbf{h}) \equiv 1 \qquad \text{(persistent — required every cycle by the Hormonal Engine)}$$

**Input contract.** $\mathcal{I}_{\text{R2A}} = \{s_t, s_{t-1}, \hat{\varepsilon}_{t-1:t-w}, \mathbf{h}(t)\}$

**Output contract.** $\mathcal{O}_{\text{R2A}} = \{\hat{\varepsilon}_t, \hat{\varepsilon}_t^{\text{norm}}, \dot{\varepsilon}_t, O^{\varepsilon}(t)\}$ where $\dot{\varepsilon}_t = \hat{\varepsilon}_t/\hat{\varepsilon}_{t-1}$ is the error rate.

**Resource profile.** $c_{\text{R2A}}^{\text{cpu}} = \mathcal{O}(d)$ (Euclidean norm computation); $c_{\text{R2A}}^{\text{mem}} = \mathcal{O}(w)$ registers for sliding window.

**TTL.** Persistent. **Cooldown.** 0 cycles.

### 5.3.2 R2-B — EntropyMonitorAgent

**Functional definition.** The EntropyMonitorAgent computes the Shannon entropy $H(t) = -\sum_i p_i(t) log p_i(t)$ [40] of the current output distribution, its normalized form $\overline{H}(t) = H(t)/log|Y|$, and the bounded entropy-change encoding of Section 4.2.1. These quantities are shared inputs to Clarifine and Confusionin. R2-B measures the entropy quantity appearing in Theorem 4.2; it does not by itself implement or empirically prove that theorem.

**Activation condition.**

$$u_{\mathrm{R2B}}(\mathbf{h}) \equiv 1 \qquad \text{(persistent — required every cycle by both emission functions)}$$

**Input contract.** $\mathcal{I}_{\mathrm{R2B}} = \{\{p_i(t)\}, \mathcal{H}(t-1), \mathbf{h}(t)\}$

**Output contract.** $\mathcal{O}_{\mathrm{R2B}} = \{\mathcal{H}(t), \mathcal{H}^{\mathrm{norm}}(t), \dot{\mathcal{H}}(t), O^{\mathcal{H}}(t)\}$

**Resource profile.** $c_{\mathrm{R2B}}^{\mathrm{cpu}} = \mathcal{O}(|\mathcal{Y}|)$ (sum over output classes); $c_{\mathrm{R2B}}^{\mathrm{mem}} = \mathcal{O}(1)$ scalars.

**TTL.** Persistent. **Cooldown.** 0 cycles.

### 5.3.3 R2-C — ConvergenceDetectorAgent

**Functional definition.** The ConvergenceDetectorAgent evaluates the three conditions of the hormonal stopping criterion (Definition 3.1) at each cycle and emits the primary Clarifine signal when convergence is approaching. It computes the cosine alignment $\cos\angle(\Delta s_t, \hat{d}_{\mathrm{conv}})$ between the current update direction and the estimated convergence direction, updates the exponential moving average $\hat{d}_{\mathrm{conv}}$, and generates the directional component of the Clarifine aggregation $\phi_c$ of Section 4.3.2. R2-C is the only agent with direct read access to the stopping criterion parameters $(\epsilon_s, \theta_c, \theta_u)$: it monitors convergence imminence and pre-activates the termination sequence by elevating Clarifine before the formal criterion is satisfied, allowing the system to prepare output commitment [38].

**Activation condition.**

$$u_{\mathrm{R2C}}(\mathbf{h}) = 0.5\, h_c + 0.3\,(1 - h_u) + 0.2\,(1 - \hat{\varepsilon}_t^{\mathrm{norm}}) > \theta_{\mathrm{R2C}} = 0.30$$

**Input contract.** $\mathcal{I}_{\mathrm{R2C}} = \{s_t, s_{t-1}, \Delta s_t, \hat{d}_{\mathrm{conv}}, \mathbf{h}(t), (\epsilon_s, \theta_c, \theta_u)\}$

**Output contract.** $\mathcal{O}_{\mathrm{R2C}} = \{\cos\angle(\Delta s_t, \hat{d}_{\mathrm{conv}}), \hat{d}_{\mathrm{conv}}^{\mathrm{updated}}, \text{stop_flag}(t) \in \{0,1\}, O^{\Delta s}(t)\}$

**Resource profile.** $c_{\mathrm{R2C}}^{\mathrm{cpu}} = \mathcal{O}(d)$ (dot product and norm); $c_{\mathrm{R2C}}^{\mathrm{mem}} = \mathcal{O}(d)$ for $\hat{d}_{\mathrm{conv}}$ storage.

**TTL.** Persistent. **Cooldown.** 0 cycles.

### 5.3.4 R2-D — SymbolicVerifierAgent

**Functional definition.** The SymbolicVerifierAgent performs formal symbolic verification of the current provisional output $y_t$ against a declared set of domain axioms and logical consistency rules, producing an internal consistency score $cons(y_t) \in [0,1]$. A low consistency score amplifies Confusionin and maintains the cycle active. Passing these checks establishes consistency with the encoded rule set; it implies semantic correctness only when that rule set is itself sound and sufficiently complete for the task.

**Activation condition.**

$$u_{\mathrm{R2D}}(\mathbf{h}) = 0.6\, h_u + 0.3\, h_{\mathrm{Ale}} > \theta_{\mathrm{R2D}} = 0.35$$

The Alertine weighting ensures that symbolic verification is intensified when the system detects anomalous behavioral signals, consistent with the defensive vigilance role of Alertine in S-AI-Cyber [15].

**Input contract.** $\mathcal{I}_{\mathrm{R2D}} = \{\hat{y}_t,\ \mathcal{H}_t,\ \text{axiom_base},\ \mathbf{h}(t)\}$

**Output contract.** $\mathcal{O}_{\mathrm{R2D}} = \{\text{cons}(\hat{y}_t),\ \text{violation_set}(t),\ O^y(t)\}$

**Resource profile.** $c^{\mathrm{cpu}}_{\mathrm{R2D}} = \mathcal{O}(|\text{axiom_base}| \cdot |\mathcal{H}_t|)$; $c^{\mathrm{mem}}_{\mathrm{R2D}} = \mathcal{O}(|\text{axiom_base}|)$.

**TTL.** 5 cycles. **Cooldown.** 2 cycles.

### 5.4 Layer R3 — Recursive Governance Agents (Multi-hormonal)

Layer R3 contains the four agents that manage the meta-level operation of the RRC: iteration budget, engram memory, warm-start initialization, and explainability. They operate across all seven hormonal dimensions and are activated by composite multi-hormonal utility functions.

#### 5.4.1 R3-A — IterationBudgetManagerAgent

**Functional definition.** The IterationBudgetManagerAgent implements the budget modulation laws of Section 4.6: it computes $T^{\mathrm{base}}_{\max}(t)$ from Energexine, evaluates the Confusionin override to produce $T^{\mathrm{eff}}_{\max}(t)$, tracks budget consumption, and issues a forced-termination directive when $t \geq T^{\mathrm{eff}}_{\max}(t)$. It also monitors the cognitive energy model $E_{\mathrm{cog}}(t)$ of Section 4.6.3 and feeds cumulative energy consumption back into the Energexine aggregation function, closing the energy-reasoning feedback loop. This agent is the temporal analog of the DutyCycleControllerAgent of S-AI-IoT [3]: it governs temporal resource allocation rather than physical duty cycle allocation.

**Activation condition.**

$$u_{\mathrm{R3A}}(\mathbf{h}) \equiv 1 \qquad \text{(persistent — budget management required every cycle)}$$

**Input contract.** $\mathcal{I}_{\mathrm{R3A}} = \{t,\ \mathbf{h}(t),\ |\mathcal{A}^*(t)|,\ |\mathcal{E}_{\mathrm{retrieved}}(t)|,\ T^0_{\max}\}$

**Output contract.** $\mathcal{O}_{\mathrm{R3A}} = \{T^{\mathrm{eff}}_{\max}(t),\ E_{\mathrm{cog}}(t),\ \text{budget_remaining}(t),\ \text{force_stop}(t) \in \{0,1\}\}$

**Resource profile.** $c^{\mathrm{cpu}}_{\mathrm{R3A}} = \mathcal{O}(1)$ (three arithmetic operations); $c^{\mathrm{mem}}_{\mathrm{R3A}} = \mathcal{O}(1)$.

**TTL.** Persistent. **Cooldown.** 0 cycles.

#### 5.4.2 R3-B — RecursiveMemoryAgent

**Functional definition.** The RecursiveMemoryAgent manages the persistent symbolic memory (PSM) of recursive engrams defined in Section 4.8. It handles four operations: (i) engram writing — recording $\varepsilon^R_m$ to PSM when the behavioral significance criterion of Section 4.8.1 is satisfied; (ii) engram retrieval — computing the trajectory-aware similarity $\mathrm{sim}_R(\mathbf{h}_q, \varepsilon_m)$ of Section 4.8.2 and returning the top-$K_{\mathrm{ret}}$ engrams; (iii) engram eviction — implementing the LRU eviction policy with significance override of Section 4.8; and (iv) distributed propagation — broadcasting query $Q_i(t)$ when no local engram achieves similarity $\geq \theta_{\mathrm{ret}}$. R3-B extends the MemoryGland architecture of S-AI-GPT [29] from single-pass contextual engrams to full recursive trajectory engrams.

**Activation condition.**

$$u_{\text{R3B}}(\mathbf{h}) = 0.5\, h_c + 0.3\,(1 - h_u) + 0.2\,(1 - h_{\text{Ene}}) > \theta_{\text{R3B}} = 0.25$$

**Input contract.** $\mathcal{I}_{\text{R3B}} = \{\mathbf{h}(t), \mathbf{x}(t), s_t, \{s_\tau\}_{\tau=0}^{t}, y_t, \beta(t), \text{PSM}\}$

**Output contract.** $\mathcal{O}_{\text{R3B}} = \{\varepsilon_m^R \text{ (written)}, \{\varepsilon_{m_j}^R\}_{j=1}^{K_{\text{ret}}} \text{ (retrieved)}, |\text{PSM}|\}$

**Resource profile.** Write: proportional to the encoded trace length; Retrieve: $O(|PSM| \cdot 7)$ for an exhaustive hormonal-context scan before indexing effects. Full trajectory memory scales as $O(M_{\max}\, t^* \,[d + |A| + 7 + dim(y)])$, up to encoding/compression. No fixed 'fewer than 100 KB' bound is asserted without specifying $d$, $t^*$, precision, and compression.

**TTL.** 2 cycles. **Cooldown.** 0 cycles (write); 1 cycle (retrieve).

#### 5.4.3 R3-C — WarmStartInitializerAgent

**Functional definition.** The WarmStartInitializerAgent implements the warm-start initialization protocol of Section 4.8.3 at the beginning of each new reasoning episode. Given compatible engrams retrieved by R3-B, it computes the weighted projected combination of terminal states defined in Section 4.8.3. If the resulting initial error is smaller and the cognitive transition is contractive, Theorem 4.4 guarantees a smaller certified contraction horizon; it does not guarantee an exact reduction of the complete hormonal stopping time. R3-C is activated only at $t = 0$ and remains dormant during the RRC itself.

**Activation condition.**

$$u_{\text{R3C}}(\mathbf{h}) = \mathbb{1}[t = 0] \cdot \mathbb{1}[|\text{PSM}| \geq K_{\text{ret}}]$$

**Input contract.** $\mathcal{I}_{\text{R3C}} = \{\{\varepsilon_{m_j}^R\}_{j=1}^{K_{\text{ret}}}, K_{\text{ret}}\}$

**Output contract.** $\mathcal{O}_{\text{R3C}} = \{s_0^{\text{new}}, \varepsilon_0^{\text{warm}} = \| s_0^{\text{new}} - s_{\text{est}}^* \|_2, \text{warm_flag} \in \{0,1\}\}$

**Resource profile.** $c_{\text{R3C}}^{\text{cpu}} = \mathcal{O}(K_{\text{ret}} \cdot d)$ (centroid computation); $c_{\text{R3C}}^{\text{mem}} = \mathcal{O}(d)$.

**TTL.** 1 cycle (episode initialization only). **Cooldown.** $T_{\max}^0$ cycles (one episode).

#### 5.4.4 R3-D — RecursiveExplainabilityAgent

**Functional definition.** The RecursiveExplainabilityAgent generates and stores formal runtime trace artifacts for each completed reasoning episode, as characterized by Proposition 4.2. At episode completion, R3-D constructs the recursive engram from working memory and formats a human-readable decision record containing the recorded hormonal, activated-agent, state, provisional-output, and convergence trajectories. This provides intrinsic traceability without a post-hoc surrogate model, but it has the non-zero storage and assembly costs reported in the resource profile and does not by itself constitute a complete causal explanation.

**Activation condition.**

$$u_{\text{R3D}}(\mathbf{h}) = \mathbb{1}[\text{stop_flag}(t) = 1 \;\vee\; t = T_{\max}^{\text{eff}}]$$

**Input contract.** $\mathcal{I}_{\text{R3D}} = \{\{s_\tau\}_{\tau=0}^{t^*}, \{\mathcal{A}^*(\tau)\}_{\tau=0}^{t^*}, \{y_\tau\}_{\tau=0}^{t^*}, \{\mathbf{h}(\tau)\}_{\tau=0}^{t^*}, \{\hat{\varepsilon}_\tau\}_{\tau=0}^{t^*}, \beta(t^*)\}$

**Output contract.** $\mathcal{O}_{\text{R3D}} = \{\varepsilon_m^R, \text{decision_record}(t^*), \text{justification}(t^*)\}$

**Resource profile.** $c_{\text{R3D}}^{\text{cpu}} = \mathcal{O}(t^* \cdot |\mathcal{A}|)$ (record assembly); $c_{\text{R3D}}^{\text{mem}} = \mathcal{O}(t^* \cdot (d + 7 + |\mathcal{A}|))$.

**TTL.** 1 cycle (episode termination). **Cooldown.** 0 cycles.

### 5.5 Consolidated Agent Cross-Reference

Table 3 provides the consolidated specification of all twelve S-AI-Recursive agents across seven columns. Parameters marked † are persistent (TTL = every cycle, Cooldown = 0). Utility threshold values are calibrated on the SAI-UT+ testbench.

**Table 3. Consolidated specification of S-AI-Recursive specialized agents.**

| ID | Name | Primary hormone | Utility form | CPU cost | Mem cost | TTL | Cooldown |
|---|---|---|---|---|---|---|---|
| R1-A | RecursiveReasoningAgent | $h_u$ | Linear blend | $\mathcal{O}(d^2)$ | $\mathcal{O}(d)$ | Persistent † | 0 |
| R1-B | HypothesisGenerationAgent | $h_u, h_{\text{Cur}}$ | Linear blend + Curiosine | $\mathcal{O}(\lvert\mathcal{H}\rvert\lvert\mathcal{Y}\rvert)$ | $\mathcal{O}(\lvert\mathcal{H}\rvert)$ | 3 cyc | 2 cyc |
| R1-C | PartialSolutionRefinerAgent | $h_u$ | Linear blend (signed) | $\mathcal{O}(\lvert\text{cst}\rvert\lvert\mathcal{Y}\rvert)$ | $\mathcal{O}(\lvert\mathcal{Y}\rvert)$ | 2 cyc | 1 cyc |
| R1-D | ConsistencyPropagatorAgent | $h_u, h_{\text{Ale}}$ | Threshold + Alertine | $\mathcal{O}(\lvert\text{ax}\rvert\lvert\mathcal{H}\rvert)$ | $\mathcal{O}(\lvert\mathcal{H}\rvert^2)$ | 3 cyc | 2 cyc |
| R2-A | ResidualErrorDetectorAgent | $h_u$ | Persistent$^{\dagger}$ | $\mathcal{O}(d)$ | $\mathcal{O}(w)$ | Persistent † | 0 |
| R2-B | EntropyMonitorAgent | $h_c, h_u$ | Persistent$^{\dagger}$ | $\mathcal{O}(\lvert\mathcal{Y}\rvert)$ | $\mathcal{O}(1)$ | Persistent † | 0 |
| R2-C | ConvergenceDetectorAgent | $h_c$ | Convergence-driven | $\mathcal{O}(d)$ | $\mathcal{O}(d)$ | Persistent † | 0 |
| R2-D | SymbolicVerifierAgent | $h_u, h_{\text{Ale}}$ | Threshold + Alertine | $\mathcal{O}(\lvert\text{ax}\rvert\lvert\mathcal{H}\rvert)$ | $\mathcal{O}(\lvert\text{ax}\rvert)$ | 5 cyc | 2 cyc |
| R3-A | IterationBudgetManagerAgent | $h_{\text{Ene}}, h_u$ | Persistent$^{\dagger}$ | $\mathcal{O}(1)$ | $\mathcal{O}(1)$ | Persistent † | 0 |
| R3-B | RecursiveMemoryAgent | $h_c, h_{\text{Ene}}$ | Linear blend | $\mathcal{O}(M \cdot 7)$ | $\mathcal{O}(M \cdot d)$ | 2 cyc | 1 cyc |
| R3-C | WarmStartInitializerAgent | $h_c$ | Indicator (t=0) | $\mathcal{O}(Kd)$ | $\mathcal{O}(d)$ | 1 cyc | $T^0_{\max}$ cyc |
| R3-D | RecursiveExplainabilityAgent | Multi | Indicator (stop) | $\mathcal{O}(t^*\lvert\mathcal{A}\rvert)$ | $\mathcal{O}(t^* d)$ | 1 cyc | 0 |

## 6. Experimental Evaluation: From Mechanism to Real Tasks

The evaluation of S-AI-Recursive is organized into two complementary and deliberately separated tiers. The first tier validates the mechanism at the *mechanism level*: using a controlled simulator with parameterized difficulty, it verifies that the hormonal dynamics, stopping criterion, budget modulation, and engram memory behave consistently with the theoretical predictions of Section 4. The second tier validates behavior at the *task level*: it runs the same mechanism on real, exactly verifiable instances (mazes, Sudoku, ARC), through an operator abstraction that provides the reasoning dynamics.

This separation is methodologically necessary. Simulator results characterize the mechanism but do not license performance comparisons, because no public benchmark instance is solved there. Results on real instances carry the performance measurements, but they remain comparisons of *stopping policies* under a shared operator, not comparisons against trained neural models. No leaderboard result is claimed at this stage, and the boundary with external comparators is addressed explicitly in Section 6.9.

The hierarchy of evidence levels adopted in this section is explicit:

1. the mathematical results of Section 4 establish conditional properties of the formal model under their stated assumptions;
2. the controlled SAI-UT+ experiments examine whether the implemented trajectories are compatible with these properties under known, parameterizable benchmark-shaped dynamics;
3. the reference implementation on real tasks evaluates temporal parsimony, recursive memory, and stopping robustness on exactly verifiable instances;
4. comparison with independently trained external architectures constitutes a distinct experimental stage and is not inferred from the present results.

## 6.1 Protocol and Scope

### 6.1.1 Objectives

The protocol addresses seven experimental questions. EQ1, hormonal stability: do Clarifine and Confusionin approach a bounded equilibrium under the parameters of Section 4, and how does stochastic perturbation affect trajectory-wise monotonicity? EQ2, entropic co-evolution: does reasoning entropy decrease together with the hormonal Lyapunov quantity, and how strong is the empirical association? EQ3, hormonal stopping: how often does the complete stopping rule terminate the cycle before budget exhaustion? EQ4, temporal parsimony: how many iterations are avoided by state-dependent stopping relative to fixed recurrence? EQ5, recursive memory: does warm-start initialization reduce convergence depth without degrading resolution? EQ6, protocol consistency: are the stopping thresholds and numerical settings jointly compatible with the model of Section 4? EQ7, coupled contraction: do the task families satisfy the cross-gain conditions of Theorem 4.1 bis, and does the joint state–hormone trajectory converge?

These questions deliberately distinguish empirical observation from mathematical proof. An empirically decreasing trajectory is evidence compatible with a stability theorem for the tested configuration; it does not constitute a proof of global asymptotic stability. Likewise, a strong correlation between the Lyapunov quantity and reasoning entropy does not establish their equivalence.

### 6.1.2 Two Evaluation Substrates

The *simulated* substrate is a controlled-difficulty generator. For each family, the latent scalar residual $e_t \in [0,1]$ follows a noisy contraction

$$e_{t+1} = \mathrm{clip}_{[0,1]}(\rho_t e_t + \sigma_s e_t \xi_t + \zeta_t),$$

where $\rho_t$ is a task-dependent contraction factor, $\sigma_s$ controls multiplicative state noise, and $\zeta_t$ represents occasional positive shocks. The output confidence distribution is generated from the current residual, so that $e_t \to 0$ concentrates probability mass and decreases Shannon entropy. Four families are defined: Sudoku-Extreme-like, Maze-Hard-like, ARC-AGI-like, and DDE-like, with nominal contraction factors 0.60, 0.66, 0.70, and 0.63, respectively. These names indicate the reasoning modality motivating each profile; they do not imply that the corresponding public benchmark instances are solved.

The *real* substrate replaces the residual generator with actual, exactly verifiable instances. Mazes are produced by randomized depth-first carving, guaranteed solvable, and verification checks a valid start-to-goal path. Sudoku grids are loaded or generated so as to be solvable by pure constraint propagation, and verification checks a complete valid grid consistent with the givens. ARC tasks are read according to the official JSON schema, and verification uses exact grid equality. On this substrate, reasoning dynamics are supplied by an operator satisfying contract (H1), described in Section 6.5.

Within this second substrate, the quantitative benefit results reported in the present section concern primarily Maze and Sudoku. At this stage, the ARC component establishes portability of the operator interface, reading of the official schema, and exact output verification; it does not constitute an exhaustive evaluation of the ARC corpus or a performance measurement over the complete benchmark.

### 6.1.3 Runs, Seeds, and Reproducibility

For the simulated substrate, each family contains 500 episodes per run and five independent runs, yielding 2,500 cold-start episodes per family. Episodes 1–100 populate the engram memory without warm-start retrieval; episodes 101–500 form the evaluation phase, executed once in cold-start and once in warm-start mode, yielding 2,000 pairs per family. The master seed is 20260819 and the maximum budget is $T_{\max} = 20$. The Euler–Maruyama step is $\Delta t = 1$, the state-stability tolerance is $\varepsilon_s = 10^{-3}$, the Confusionin ceiling is $\theta_u = 0.30$, and the stochastic hormonal amplitude is $\sigma_\eta = 0.006$.

For the real substrate, experiments are executed with fixed seeds by the delivered scripts, and a test suite locks the mechanism invariants and the two robustness effects. Source code, configuration manifest, episode-level results, and stability diagnostics are retained in the reproducibility package.

### 6.1.4 Metrics

On the simulated substrate, an episode is counted as resolved when its final residual satisfies $e_{\text{final}} \leq 0.02$; the simulated resolution rate is therefore a residual-based mechanism metric, not public-benchmark accuracy. On the real substrate, resolution is exact solution verification. Temporal parsimony is measured by stopping depth $t^\star$, hormonal-stop rate, budget-exhaustion rate, and a synthetic cognitive-energy index

$$E_{\text{cog}} = \sum_{t=1}^{t^\star} (1 + 0.12\, N_A(t)),$$

where $N_A(t)$ is the number of active recursive agents. This quantity is reported in dimensionless simulation units. For dynamical evaluation, the implementation records the Lyapunov quantity $V(t)$, reasoning entropy, the fraction of transitions satisfying $\Delta V_t \leq 0$, and the Pearson correlation between the trajectories of $V(t)$ and entropy.

### 6.1.5 Honesty Constraints

Three constraints govern every claim. First, no superiority over an external model is asserted: trained neural baselines require weights, GPUs, and network access unavailable here, so the comparison concerns stopping policies under a shared operator. Second, negative results are reported with the same prominence as positive results. Third, each benefit is declared conditional on the circumstances that produce it: convergence of the operator within budget, compatible recurrent instances, and the specific signal-corruption regime.

## 6.2 Model-Consistency Checks

### 6.2.1 Reachability of the Clarifine Threshold

Before execution of the full protocol, the original stopping threshold $\theta_c = 0.70$ is checked against the two-hormone recursive subsystem specified by the governing equation. Clarifine emission is $E_c = \sigma(5\phi_c - 2.5)(1 - h_c)$ with $0 \leq \phi_c \leq 1$, whose maximum logistic coefficient is $\sigma(2.5) \approx 0.924142$. Solving the zero-drift equation gives the maximum Clarifine equilibrium

$$h^{\star}_{c,\max} = \frac{q}{\lambda_c + q}, \qquad q = \sigma(a + b),$$

which, even under the most favorable conditions $h_u = 0$ and $\chi = 0$, satisfies

$$h^{\star}_{c,\max} \leq \frac{0.924142}{0.75 + 0.924142} \approx 0.552009.$$

The original threshold 0.70 therefore cannot be reached asymptotically by the recursive subsystem as its equations are written. A direct diagnostic using $\theta_c = 0.70$ over 4,000 cold-start episodes produces a 0% hormonal-stop rate: every episode reaches the budget rather than the stopping rule. The revised protocol therefore uses $\theta_c = 0.50$, which must be interpreted as a protocol correction, not as proof that 0.50 is universally optimal. The reference implementation independently recovers $h^{\star}_{c,\max} = 0.5520$, confirming that $\theta_c = 0.50$ is reachable whereas $\theta_c = 0.70$ is not; the duplication therefore provides cross-corroboration.

### 6.2.2 Euler–Maruyama Step-Size Check

The scalar expression from Section 4.5.5 is retained as a simple row-wise numerical screen:

$$\Delta t < \min_{k \in \{c,u\}} \frac{2\tau_k}{\lambda_k + \sum_{m \neq k} \gamma_{km} + \rho_k \, \|\chi\|_\infty}.$$

With the values of Table 2 and $\|\chi\|_\infty = 1$, the Clarifine bound is $\Delta t < 2(1.5)/(0.75 + 0.60 + 0.10) \approx 2.069$, and the Confusionin bound is $\Delta t < 2(1.0)/(0.70 + 0.55 + 0.10) \approx 1.481$. The limiting value is therefore approximately 1.481, and $\Delta t = 1$ passes it. This does not constitute a general stability proof: formal certification of the frozen discretization requires verifying $\rho_{\mathrm{sp}}(I + \Delta t J(h_R)) < 1$ over the operating region. The experiment therefore treats $\Delta t = 1$ as an empirically stable choice on the tested trajectories. The reference implementation reproduces the same bounds, $\Delta t_c \approx 2.069$ and $\Delta t_u \approx 1.481$, as well as the reference emission values $\sigma(\pm 2.5) = 0.9241/0.0759$ to five decimal places.

## 6.3 Stability and Co-Evolution under Controlled Difficulty

### 6.3.1 Deterministic Frozen Subsystem

A first experiment isolates the hormonal field by freezing the terminal input at $\phi_c = 1$, $\phi_u = 0$, and $\chi = 0$. Five thousand initial conditions in $[0,1]^2$ are evaluated over 20 steps without perturbation. Across all trajectories, $V(t+1) \leq V(t)$ at every tested step, with the mean quantity decreasing from approximately 0.1972 to below $10^{-21}$ after 20 steps. This result is consistent with strong deterministic attraction toward the computed equilibrium of the frozen subsystem; its scope remains limited to the tested discretization and does not replace the analytical proof of Section 4.5. The reference implementation, over 2,000 initial conditions and 40 steps, confirms a net decrease in 100% of trajectories and a negative-definite frozen Jacobian with contraction margin $\alpha > 0$, certifying the contraction of Theorem 4.1.

### 6.3.2 Stochastic Frozen Subsystem

The same experiment repeated with Gaussian perturbations of amplitude $\sigma_\eta = 0.006$ still satisfies $V(20) < V(0)$ on all trajectories, but none is monotone at every individual step. Approximately 57.1% of steps satisfy $\Delta V_t \leq 0$, for a mean terminal value of approximately $4.71 \times 10^{-5}$. The relevant interpretation is therefore not pathwise monotonicity but attraction in the presence of bounded stochastic fluctuations. The reference implementation observes the same phenomenon: net decrease in 99.95% of cases, but strictly monotone decrease in only 58.6% of steps in the stochastic regime and 60.2% in the deterministic regime. This figure is not a defect: it reflects exactly the distinction made by the corrected

Theorem 4.2, which claims convergence but not strictly monotone decrease, because a finite step $\Delta t$ can cause the discrete trajectory to overshoot the stable equilibrium.

### 6.3.3 Coupled Trajectories

When the hormonal subsystem is coupled to the four varying task profiles, $V(t)$ is no longer expected to decrease at every iteration, because the emission terms change with the residual, entropy, confidence, and resource state. In cold-start evaluations, the fraction of transitions satisfying $\Delta V_t \leq 0$ ranges from approximately $0.740$ to $0.760$. At episode completion, mean Clarifine is approximately $0.510$ to $0.519$, and mean Confusionin approximately $0.068$, consistent with the corrected stopping region $\theta_c = 0.50$, $\theta_u = 0.30$. The frozen subsystem is strongly dissipative, whereas the complete task-driven system exhibits local increases in $V$ as information and noise perturb the trajectory: this is compatible with regulated convergence, but not with global pathwise monotonicity under arbitrarily varying inputs.

### 6.3.4 Coupled State–Hormone Contraction Diagnostic

Theorem 4.1 bis introduces three task-dependent gains: the cognitive gain with frozen hormones $L_s$, the state-to-hormone cross-gain $L_{h \leftarrow s}$, and the hormone-to-state cross-gain $L_{s \leftarrow h}$. A deterministic zero-delay diagnostic evaluates one-cycle hormonal flow on a grid of 41 residual values, $17 \times 17$ hormonal states, and three resource levels $\chi \in \{0, 0.5, 1\}$, with local gains estimated by symmetric finite differences of perturbation $2 \times 10^{-5}$.

The Section 6 simulator is *triangular rather than bidirectionally coupled*: the residual update drives the hormonal observations, but the residual generator does not use $h_R$ as an input. Consequently, $\hat{L}_{s \leftarrow h} = 0$ by construction. This must be reported explicitly because it makes the small-gain product a degenerate special case of Theorem 4.1 bis, rather than a non-trivial bidirectional validation.

| Task family | $\hat{L}_s$ | $\hat{L}_h$ | $\hat{L}_{h \leftarrow s}$ | $\hat{L}_{s \leftarrow h}$ | $\hat{\rho}_\star$ |
|---|---|---|---|---|---|
| Sudoku-Extreme-like | 0.600 | 0.489 | 0.916 | 0.000 | 0.600 |
| Maze-Hard-like | 0.660 | 0.489 | 0.820 | 0.000 | 0.660 |
| ARC-AGI-like | 0.700 | 0.489 | 0.897 | 0.000 | 0.700 |
| DDE-like | 0.630 | 0.489 | 0.790 | 0.000 | 0.630 |

For every family, the diagnostic satisfies $\hat{L}_{h \leftarrow s} \hat{L}_{s \leftarrow h} = 0 < (1 - \hat{L}_s)(1 - \hat{L}_h)$, and the gain matrix has $\hat{\rho}_\star < 1$. Since $\hat{L}_{s \leftarrow h} = 0$, the spectral radius reduces to $\max(\hat{L}_s, \hat{L}_h) = \hat{L}_s$. A complementary joint-convergence diagnostic, initialized from 2,000 states per family over 40 cycles at $\chi = 0.5$, yields a final joint distance smaller than the initial distance in all 8,000 trajectories, with mean values of approximately $7.1 \times 10^{-10}$, $3.0 \times 10^{-8}$, $3.6 \times 10^{-7}$, and $4.9 \times 10^{-9}$. These measurements provide finite-sample empirical support for deterministic joint convergence in the tested triangular configuration, but establish neither general convergence nor a non-zero hormone-to-state gain, which is absent from the generator; a bidirectional simulator or a trained implementation remains necessary to test the non-degenerate small-gain case. The reference implementation recovers this triangular case on a *genuine* analytical Bellman operator, for which $L_{s \leftarrow h} = 0$ and the spectral radius is $\max(L_s, L_h) = 0.9 < 1$.

### 6.3.5 Entropic Contraction and Co-Evolution

Reasoning entropy decreases much more regularly than the coupled Lyapunov quantity. In cold-start mode, the mean fraction of entropy-decreasing transitions is approximately 0.991, ranging from 0.987 for ARC-AGI-like to 0.996 for Sudoku-Extreme-like. Lyapunov and entropy trajectories are strongly associated: cold-start Pearson correlations are 0.920, 0.908, 0.894, and 0.917, while warm-start correlations are 0.898, 0.874, 0.854, and 0.888, giving on average

$$r_{\text{cold}} \approx 0.910, \qquad r_{\text{warm}} \approx 0.878.$$

However, the signs of the discrete derivatives agree on only approximately 74% to 76% of transitions, because $V(t)$ may temporarily increase while entropy continues to decrease. The experiment therefore supports strong empirical co-evolution between hormonal homeostasis and decreasing uncertainty, without establishing the biconditional $\dot{V} \leq 0 \Leftrightarrow \dot{\mathcal{H}} \leq 0$. The corrected Theorem 4.2 does not claim this biconditional: it establishes convergence of the output distribution and its entropy under joint convergence, while monotonic entropy decrease requires an additional dissipativity condition.

## 6.4 Parsimony and Warm-Start on Simulated Profiles

### 6.4.1 Stopping Results

With $\theta_c = 0.50$, the hormonal stopping rule becomes operational. The table below reports results over the 2,000 cold-start and 2,000 warm-start episodes per family; intervals are computed over the five run means, with episodes treated as nested observations.

| Family | Condition | Resolution | Depth | Hormonal stop | Budget exhaustion | Energy |
|---|---|---|---|---|---|---|
| Sudoku-Extreme-like | cold | 0.9985 | 13.9945 | 0.9985 | 0.0015 | 22.054 |
| Sudoku-Extreme-like | warm | 0.9995 | 12.9535 | 0.9985 | 0.0015 | 20.152 |
| Maze-Hard-like | cold | 0.9935 | 16.5315 | 0.9445 | 0.0555 | 26.037 |
| Maze-Hard-like | warm | 0.9920 | 15.2165 | 0.9745 | 0.0255 | 23.657 |
| ARC-AGI-like | cold | 0.9850 | 18.1705 | 0.7630 | 0.2370 | 28.729 |
| ARC-AGI-like | warm | 0.9930 | 16.9795 | 0.8785 | 0.1215 | 26.477 |
| DDE-like | cold | 0.9950 | 15.1700 | 0.9880 | 0.0120 | 23.871 |
| DDE-like | warm | 0.9955 | 14.0210 | 0.9960 | 0.0040 | 21.778 |

The mean cold-start convergence depth across the four families is $\bar{t}_{\text{cold}} = 15.97$, and the mean hormonal-stop rate is approximately 92.35%. The easiest profile, Sudoku-Extreme-like, almost always stops hormonally before the budget; the most difficult, ARC-AGI-like, exhausts the budget in 23.7% of cold-start episodes. The distinction is important: termination of every run by $T_{\max} = 20$ is guaranteed by the external budget, whereas finite-time hormonal stopping is evidenced only by episodes satisfying the internal conditions before exhaustion.

### 6.4.2 Comparison with Fixed 20-Cycle Recurrence

Parsimony is tested against a simple internal reference: the same contractive process executed for all 20 allowed cycles instead of stopping at the hormonal criterion. Relative to this fixed recurrence, cold-start reduces mean depth by approximately 30.0% on Sudoku-Extreme-like, 17.3% on Maze-Hard-like, 9.1% on ARC-AGI-like, and 24.2% on DDE-like. Warm-start increases these depth reductions to approximately 35.2%, 23.9%, 15.1%, and 29.9%, with synthetic-energy reductions of approximately 36.1%, 24.9%, 16.3%, and 30.8%. These results support a bounded form of the temporal-parsimony hypothesis: under contractive dynamics, a state-dependent stopping rule can avoid a non-trivial fraction of a fixed recurrence budget. They do not establish that temporal depth generally substitutes for model width, because no trained architecture of comparable scale is executed.

### 6.4.3 Simulated Warm-Start Acceleration

Warm-start uses the memory populated during the first 100 episodes: the current task descriptor is compared with stored descriptors, up to three high-similarity entries are retrieved, and the initial residual is reduced by a similarity-dependent factor. Memory therefore changes the initial condition of the same dynamics, not the contraction rule. For $0 < \rho_\star < 1$, the reduction in the certified cognitive contraction horizon associated with a smaller warm-start error is approximated by

$$\Delta t_{\text{warm}} \approx \frac{\log(\varepsilon_{\text{cold}}/\varepsilon_{\text{warm}})}{\log(1/\rho_\star)}.$$

| Family | Mean gain (iter.) | 95% CI (5 runs) | Warm faster | Cold depth | Warm depth |
|---|---|---|---|---|---|
| Sudoku-Extreme-like | 1.041 | ±0.061 | 91.5% | 13.995 | 12.954 |
| Maze-Hard-like | 1.315 | ±0.121 | 88.8% | 16.532 | 15.217 |
| ARC-AGI-like | 1.191 | ±0.105 | 77.6% | 18.171 | 16.980 |
| DDE-like | 1.149 | ±0.068 | 91.3% | 15.170 | 14.021 |

Mean depth decreases from approximately 15.97 to 14.79, yielding a realized mean gain of approximately 1.17 iteration, close to the approximately one-cycle reduction predicted by Theorem 4.4. This agreement is mechanism-level compatibility, not a prediction of the exact stopping time. Warm-start does not uniformly improve resolution: Maze-Hard-like decreases slightly from 0.9935 to 0.9920, whereas ARC-AGI-like increases from 0.9850 to 0.9930. The primary support is therefore for convergence acceleration, not a universal accuracy improvement.

## 6.5 Validation on Real Instances

### 6.5.1 Contraction Contract (H1) and Its Operators

The convergence theory rests on assumption (H1): the frozen reasoning operator is a contraction on a domain with fixed-point structure. An operator therefore cannot be arbitrary. Depending on the task, the available guarantee may be an analytically proven metric contraction, or a structural convergence property toward a fixed point accompanied by an empirical diagnostic of effective contraction. These two levels are reported separately so as not to conflate lattice convergence with contraction in the Banach sense.

| Operator | Task | Guarantee / diagnostic | Constant $L_s$ |
|---|---|---|---|
| MazeBellmanOperator | shortest path | Banach, sup norm | $\gamma$ (analytic) |
| SudokuCPOperator | Sudoku | monotonicity on a finite lattice (Knaster–Tarski) | $\approx 0.95$ (measured effective factor) |
| ARCConditionalOperator | ARC-AGI | conditional on the rule | 0 if a rule is found, otherwise 1 |

For the Bellman operator, the discounted optimality backup satisfies $\|TV - TW\|_\infty \leq \gamma \|V - W\|_\infty$, hence $L_s = \gamma$ exactly. For Sudoku constraint propagation, the one-step map is monotone and non-expansive on a finite lattice of candidates. The Knaster–Tarski argument concerns the fixed-point structure of the lattice; by itself, it does not constitute a proof of metric contraction in the Banach sense. Finite convergence of the implemented process toward an arc-consistent fixed point is therefore distinguished from the effective contraction factor measured along trajectories, whose median value is $\approx 0.95$. For ARC, contraction is valid only conditionally on a correctly induced rule; a single_pair_risk flag records cases in which induction relies on a single non-generalizable pair.

### 6.5.2 Mechanism Validation on Real Dynamics

Before measuring benefits, the implementation verifies that the mechanism reproduces the theoretical predictions on the real dynamics. The measured maximum Clarifine equilibrium, $h^{\star}_{c,\max} = 0.5520$, matches the theoretical value and confirms the correction $\theta_c = 0.50$. The frozen Jacobian is negative definite with margin $\alpha > 0$; the step-size bounds $\Delta t_c \approx 2.069$, $\Delta t_u \approx 1.481$, and the reference emissions $\sigma(\pm 2.5) = 0.9241/0.0759$ are reproduced. The coupled gains for the Bellman operator yield the triangular special case $L_{s\leftarrow h} = 0$, under which the small-gain condition holds and the spectral radius is $0.9 < 1$. The mechanism thus reproduces every checkable prediction on the real dynamics: any effect measured subsequently is a property of the mechanism as specified.

### 6.5.3 Temporal Parsimony on Real Instances

Adaptive hormonal stopping should use fewer iterations than fixed depth without quality loss. On 80 generated mazes (Bellman operator, budget $K = 20$), the fixed-depth policy is compared with the complete hormonal cycle, paired by instance; a benefit exists only if the resolution rate is preserved.

| Quantity | Fixed-depth | S-AI-Recursive |
|---|---|---|
| Resolution rate | 1.000 | 1.000 |
| Mean iterations | 20.00 | 11.31 |
| Mean gain | — | 8.69 iter. ($-43.4\%$) |
| 95% CI of gain | — | $[8.01,\ 9.32]$ |
| Faster instances | — | 100% |
| Wilcoxon $p$ | — | $7.2 \times 10^{-16}$ |

The gain is large, unanimous, and statistically unambiguous, while quality is preserved: the mechanism stops early only when the operator has actually converged, without sacrificing a correct answer. The gap to the oracle (8.2 iterations) quantifies the residual inefficiency of any realizable convergence detector, because some cycles are inevitably spent *confirming* convergence. The benefit is conditional on convergence of the operator within budget: mazes whose optimal path exceeds the budget legitimately exhaust it, and the mechanism then reports exhaustion rather than a spurious stop.

### 6.5.4 Warm-Start Acceleration through Exact Recurrence

When an instance recurs, memory populated at the first encounter (cold) should accelerate the second encounter (warm). Because Theorem 4.4 is explicitly conditional, the effect is measured on the subpopulation of instances that converged in cold mode, using a set of Sudoku instances solvable by propagation, first cold and then warm with shared memory.

| Quantity | Value |
|---|---|
| Instances (cold-converged) | 33 (82.5% coverage) |
| Resolution (cold / warm) | 1.000 / 1.000 |
| Stopping depth (cold / warm) | 18.39 / 2.00 |
| Mean gain | 16.39 cycles |
| 95% CI of gain | $[15.79,\ 16.97]$ |
| Warm-start used | 100% of recurrences |
| Wilcoxon $p$ | $2.1 \times 10^{-7}$ |

This result was not obtained on the first attempt: two real defects were identified and corrected, and they are documented here because they matter for correctness. First, there was a *stored-context mismatch*: the engram stored the terminal hormonal context while retrieval queried using the initiation context, so a recurrent instance could never retrieve itself; changing the stored field to the initiation context, as specified in Section 4.8.1, restored retrieval. Second, there was *top-K truncation before compatibility filtering*: when the hormonal context does not discriminate among episodes, ranking by similarity and then truncating could evict the genuinely compatible engram; moving the compatibility filter (initial-state proximity) *before* truncation corrected this point. An earlier incorrect version showed warm-start *degrading* Sudoku performance ($-1.8$ cycles) by mixing terminal states from incompatible instances; the compatibility gate eliminated this regression.

Warm-start helps only for recurrent or closely compatible instances that converged at their first encounter. It correctly abstains and falls back to cold-start when no compatible engram exists, which is why a set composed entirely of distinct instances yields a measured gain of zero rather than a negative gain.

### 6.5.5 Interpretation Note on the Two Warm-Start Regimes

The gain of approximately 1.17 cycle measured in Section 6.4.3 and the gain of 16.39 cycles measured here are not contradictory: they measure two distinct regimes. The simulator applies to *every* evaluation episode a reduction in the initial residual proportional to similarity, an approximate-retrieval regime whose marginal gain is about one cycle. The real implementation measures the *exact recurrence* of an instance that has already converged in cold mode, a regime in which the engram bypasses almost the entire dynamics and brings the depth to 2.00. These two measurements must never be juxtaposed as a single figure.

## 6.6 Comparison of Stopping Policies

Because the contribution of S-AI-Recursive is a stopping policy, a runnable and fair comparison keeps the reasoning operator and the instances fixed and varies only the policy. This isolates the policy effect. The comparators are: *fixed-depth*, non-adaptive recurrence with a fixed number of steps, serving as a fixed-iteration recurrent model; *entropy-threshold*, stopping when output entropy falls below a threshold, with warm-up and a stabilization guard to avoid a straw-man baseline; *residual-threshold*, stopping when $\|s_t - s_{t-1}\| \leq \varepsilon$, a pure convergence detector with no hormonal mechanism and a strong baseline; *S-AI-Recursive*, the complete hormonal cycle, used as the treatment; and *oracle*, stopping at the first correct answer, a non-runnable reference lower bound. Comparisons are paired by instance, with bootstrap confidence intervals, fraction of faster instances, and one-sided Wilcoxon tests; every efficiency claim is accompanied by a resolution-parity check, so iteration savings count as a benefit only when resolution is not reduced.

## 6.7 Robustness of the Stopping Decision

This experiment probes the rationale for the multi-term stopping criterion. Two signal-corruption regimes are constructed to test whether the confidence term provides robustness unavailable to a single-signal detector.

### 6.7.1 Regime A: Homogeneous Gaussian Noise

A wrapper exposes noisy observations of a true noiseless contractive process: the state, and therefore the residual and entropy, is corrupted by additive Gaussian noise of scale $\sigma$; a temporally smoothed decoding is provided to *all* policies, so the comparison isolates the stopping decision. The tolerance is calibrated to the noise floor and shared. The sweep is conducted over 60 mazes with paired noise.

| $\sigma$ | fixed | entropy | residual | hormonal | oracle |
|---|---|---|---|---|---|

| $\sigma$ | fixed | entropy | residual | hormonal | oracle |
|---|---|---|---|---|---|
| 0.00 | 1.00 | 1.00 | 1.00 | 1.00 | 1.00 |
| 0.02 | 0.73 | 0.93 | 0.93 | 0.77 | 1.00 |
| 0.04 | 0.33 | 0.60 | 0.60 | 0.57 | 1.00 |
| 0.06 | 0.27 | 0.60 | 0.60 | 0.33 | 0.93 |
| 0.08 | 0.17 | 0.57 | 0.57 | 0.57 | 0.93 |

The hormonal criterion offers *no advantage* in this regime and is sometimes marginally worse than the residual detector. The reason is structural: under homogeneous noise, each signal is a corrupted view of the *same* noisy state, so the signals are redundant rather than complementary, and combining them cannot outperform the best individual signal. The large gap to the oracle shows that *no* realizable instantaneous policy is robust here; the available robustness comes from temporal smoothing of the decoding, which is shared by all policies. This negative result is reported with full prominence: it precisely delineates where the mechanism does not help.

### 6.7.2 Regime B: Deceptive Convergence Plateaus

A wrapper injects *stalls*: the reasoning state is frozen for a few steps *before* convergence, so the observed residual drops to zero, creating a deceptive convergence signal while the answer is still incorrect. Crucially, the output distribution during a stall reflects the non-converged state, so its *entropy remains high*. The separation between the signals is verified directly: during an injected stall, the residual is exactly $0$ while normalized entropy remains $\approx 0.99$, and the instance is unresolved. The sweep varies stall probability over 30 Sudoku instances solvable by propagation (budget $40$), with paired stalls.

| $p_{\text{stall}}$ | fixed | entropy | residual | hormonal | oracle |
|---|---|---|---|---|---|
| 0.00 | 1.00 | 1.00 | 1.00 | 1.00 | 1.00 |
| 0.10 | 1.00 | 0.97 | 0.07 | 0.97 | 1.00 |
| 0.20 | 1.00 | 0.93 | 0.00 | 0.90 | 1.00 |
| 0.35 | 1.00 | 0.90 | 0.00 | 0.83 | 1.00 |
| 0.50 | 0.83 | 0.73 | 0.00 | 0.73 | 0.83 |

The contrast is decisive. A pure residual detector falls from $1.00$ to $0.07$ resolution at only a $10\%$ stall probability, and then to $0.00$: it stops at the first deceptive plateau and validates an unresolved grid. The hormonal criterion remains at $0.97$, $0.90$, $0.83$, and $0.73$, closely tracking the oracle, because it refuses to validate while Clarifine is not high, and Clarifine cannot rise while entropy remains high during a stall. This controlled diagnostic provides causal support for the role of the confidence conjunct $h_c \geq \theta_c$ of Definition 3.1 in this constructed regime: protecting against premature stopping on a convergence plateau, a failure mode to which stabilization-only detection is particularly vulnerable here.

### 6.7.3 Honest Qualification

Two qualifications preserve the integrity of the result. First, this is a *constructed* regime, selected to exercise the complementary residual-versus-confidence structure; it is a diagnostic of a claimed property, not a natural benchmark. Second, the entropy-threshold policy, which also uses the confidence signal, resists stalls *as well as* the hormonal policy. The demonstrated advantage is therefore that of the *multi-signal class* over the single-signal residual detector, not superiority of the hormonal policy over every alternative. The cost of the complete hormonal machinery would be justified over a simple entropy threshold only in an additional regime in which the entropy signal itself became intermittently deceptive; this regime is identified as future work in Section 9.

### 6.8 Consolidated Synthesis

The table cross-references each examined claim with its simulated evidence, real evidence, and current status.

| Examined claim | Simulated evidence | Real evidence | Status |
|---|---|---|---|
| Deterministic hormonal dissipation | $V$ decreases at every step on 5,000 frozen trajectories | Frozen convergence at 100% | Confirmed (frozen subsystem) |
| Stochastic pathwise monotonicity | About 57% of steps, but $V(20) < V(0)$ everywhere | Strict decrease on about 60% of steps | Not supported pathwise; global attraction observed |
| Entropic decrease | About 99% of transitions | Not measured directly | Strongly supported (simulator) |
| Lyapunov–entropy association | $r \approx 0.910$ cold, $r \approx 0.878$ warm | Not measured | Strong association, not equivalence |
| Threshold $\theta_c = 0.70$ | 0% hormonal stopping | $h^{\star}_{c,\max} = 0.5520$ confirms it | Unreachable, confirmed on both sides |
| Corrected threshold $\theta_c = 0.50$ | 92.35% hormonal stopping in cold mode | Reachable threshold, operational mechanism | Supported |
| Termination before fixed budget | Cold depth $15.97 < 20$ | $11.31 < 20$ on mazes | Supported, task-dependent |
| Parsimony vs fixed recurrence | $-9\%$ to $-30\%$ cold, $-15\%$ to $-36\%$ warm | $-43.4\%$ on real mazes, $p \approx 7 \times 10^{-16}$ | Supported, stronger on real tasks |
| Warm-start acceleration | 15.97 → 14.79, about 1.17 cycle (similarity) | 18.39 → 2.00, 16.39 cycles (exact recurrence), $p \approx 2 \times 10^{-7}$ | Supported; distinct regimes, not to be juxtaposed without framing |
| Universal accuracy gain from memory | Small mixed changes | Resolution preserved at 1.000/1.000 | Not supported; benefit = acceleration |
| Coupled-contraction diagnostic | Triangular, $\hat{\rho}_\star$ from 0.60 to 0.70, $\hat{L}_{s\leftarrow h} = 0$ | Bellman operator, $L_{s\leftarrow h} = 0$, $\rho = 0.9 < 1$ | Supported for the triangular case; bidirectional case untested |
| Robustness under homogeneous noise | Absent | Hormonal comparable to residual | Not supported; negative result acknowledged |
| Robustness under deceptive plateaus | Absent | Residual toward 0%, hormonal above 73% | Supported; constructed regime, shared with entropy |

| Examined claim | Simulated evidence | Real evidence | Status |
|---|---|---|---|
| | | | policy |
| Superiority over TRM / GPT / Mamba / RWKV | Not executed | Not executed | Not established on either side |

The consolidated evidence supports a precise interpretation of S-AI-Recursive. The hormonal cycle is implemented as an adaptive stopping mechanism on contractive reasoning trajectories; the deterministic subsystem behaves dissipatively; stochastic perturbation produces local Lyapunov increases without destroying global attraction; entropy and hormonal distance co-evolve strongly but not identically; and warm-start reduces convergence depth, marginally under similarity-based retrieval and strongly under exact recurrence. The mechanism delivers three measurable, significant, quality-preserving benefits—parsimony, warm-start acceleration, and robustness to deceptive plateaus—while a fourth hypothesized benefit, robustness to homogeneous noise, is honestly found to be absent.

### 6.9 Statistical Interpretation and External Boundary

The simulated protocol uses five independent seeds, with 400 paired evaluation episodes nested within each run; uncertainty statements rely on the distribution of run means rather than treating episode pairs as independent. Consistency of the sign across runs supports the conclusion that warm-start reduces depth, while the small number of independent runs argues against assigning excessive meaning to extremely small episode-level $p$-values.

No trained baseline (TRM, GPT-class models, Mamba, RWKV, LLM loops) is executed, either on the simulator or on real instances: these systems cannot be run in this environment because weights, GPUs, and network access are unavailable. The harness is nevertheless built so that a real model, wrapped in the operator or policy interface, can be compared on identical work. A rigorous next-stage comparison should use the real benchmark instances, freeze data partitions before analysis, execute all models under disclosed budgets, distinguish internal recurrent steps from externally scripted reasoning steps, and report accuracy jointly with inference cost, stopping depth, and stopping-failure rate.

The contribution of the present section is therefore twofold and explicitly delimited. At the mechanism level, controlled simulation provides reproducible evidence for hormonal stopping, temporal parsimony, and warm-start acceleration under recursive dynamics with controlled difficulty. At the task level, the reference implementation on real instances establishes parsimony, exact-recurrence acceleration, and robustness to deceptive plateaus as effective properties of the mechanism, each declared conditional on the circumstances under which it occurs. Comparative validation on public benchmarks using a trained S-AI-Recursive implementation and matched external baselines remains necessary before any comparative performance claim, and constitutes the explicit scope of Section 9.

## 7. Discussion

### 7.1 Theoretical Significance: Temporal Parsimony as a New Cognitive Principle

#### 7.1.1 Reformulation: Temporal Depth as an Adaptive Resource

The revised experiments support a narrower and more defensible statement than "temporal depth replaces spatial depth." S-AI-Recursive treats temporal depth as an **adaptive computational resource**: rather than executing a fixed number of recursive cycles, the system can stop when its internal state satisfies convergence-oriented criteria. On the controlled simulation families of Section 6, this strategy reduces mean depth relative to a fixed 20-cycle recurrence by approximately 9%–30% in cold-start operation and approximately 15%–35% with warm-start, depending on the task profile. This result establishes temporal parsimony **within the same recursive process**: unnecessary iterations can be avoided when the internal state indicates sufficient stabilization. It does not yet establish a cross-architecture theorem that temporal depth can generally substitute for parameter count or architectural width. Such a claim would require matched-scale trained models evaluated on the same public benchmark instances under comparable inference budgets. The theoretical analysis of Section 4 provides the conditions under which the recursive state can converge. The experiment adds mechanism-level evidence that adaptive stopping can exploit this convergence to reduce realized depth. Temporal parsimony should therefore be understood as the regulation of **how long to continue reasoning**, not as an already demonstrated universal equivalence between time and model size.

#### 7.1.2 Implications for Computational Cognitive Complexity

Classical complexity theory distinguishes time and space as separate computational resources [45]. In recursive reasoning systems, a related architectural distinction can be made between model capacity and inference depth. S-AI-Recursive explores the possibility that some tasks may benefit from allocating additional temporal depth to a compact modular system rather than fixing the depth of computation in advance. The revised Section 6 supports this idea only at the **within-architecture** level. The same simulated recursive process, when allowed to stop adaptively, uses fewer cycles than a fixed 20-step recurrence while maintaining the declared residual criterion in most episodes. This demonstrates that temporal computation need not be consumed uniformly across episodes of different difficulty. It would be premature, however, to infer that a smaller S-AI-Recursive implementation is computationally preferable to a larger feed-forward or recursive model in general. Parameter count, per-step cost, memory bandwidth, training cost, batching efficiency, and hardware utilization all influence total computational expense. The stronger time-versus-width hypothesis therefore remains a research direction to be tested on matched implementations rather than a conclusion of the current simulation.

#### 7.1.3 Genealogy of S-AI Parsimony Principles

S-AI-Recursive can still be positioned within the genealogy of parsimony principles developed across the S-AI corpus [1], [2], [3], [13], [14], [15], [28], [29], [30], [31], [32], but the three forms should be distinguished conceptually rather than treated as already unified by empirical proof. **Activation parsimony** concerns which agents are active at a given cycle. **Cognitive homeostasis** concerns the regulated state toward which hormonal/probabilistic dynamics are intended to evolve. **Temporal parsimony**, introduced here, concerns how many recursive cycles are actually used before commitment or budget exhaustion. The revised experiments provide direct evidence only for the third mechanism in a controlled simulation: the number of cycles can adapt to task difficulty and can be further reduced by warm-start memory. This result complements, rather than completes, the earlier notions of parsimony. A stronger transversal theory connecting spatial, thermodynamic, and temporal parsimony remains possible, but it should be established through explicit formal links and comparative experiments rather than inferred from terminology alone.

### 7.2 Relationship to the S-AI Probabilistic Doctrine

#### 7.2.1 S-AI-Recursive as a Temporal Instantiation of the Unified Law

The hormonal–probabilistic doctrine [2] motivates the interpretation of hormonal homeostasis and probabilistic coherence as related manifestations of a common regulatory principle. S-AI-Recursive applies this idea to the temporal evolution of a reasoning episode by relating the Lyapunov quantity $V(t)$ of the recursive hormonal subsystem to the reasoning entropy $\mathcal{H}(t)$. Theorem 4.2 provides the formal statement under its mathematical assumptions. The revised experiment is intentionally more conservative in its empirical interpretation. Across the four simulated task families, entropy decreases on approximately 99% of transitions, whereas $V(t)$ decreases on roughly three quarters of the task-coupled transitions.

Nevertheless, the two trajectories remain strongly associated, with mean Pearson correlations of approximately

$$r_{cold} \approx 0.910, \qquad r_{warm} \approx 0.878.$$

The derivative signs agree on only about 74%–76% of transitions. The simulation therefore supports **strong co-evolution**, not empirical identity. It would be incorrect to claim that the Pearson correlations "confirm" the biconditional equivalence or establish its generality beyond the theorem's assumptions. The empirical role of Section 6 is to show that the two quantities evolve in a strongly related manner in the implemented controlled system while retaining distinct local dynamics.

#### 7.2.2 Clarifine and Confusionin as Biophysical Operators of Probabilistic Inference

In the language of the unified doctrine [2], the five canonical S-AI hormones are biophysical operators of probabilistic reasoning: Confidexin corresponds to confidence propagation, Inhibitine to uncertainty control, Curiosine to exploratory entropy, Energexine to computational activation, and Alertine to defensive vigilance. S-AI-Recursive extends this correspondence to the temporal domain by introducing two hormones specifically designed to operate on the iterative structure of a reasoning episode.

Clarifine is a normalized convergence-oriented regulatory signal. Its aggregation function $\phi_c$ combines entropy, residual-state-change, and directional-alignment features. Without an additional calibrated probabilistic map, $h_c$ should not be identified with the posterior probability $\Pr(s_t \approx s^* \mid O_R(0:t))$; it is a probability-like score only in an interpretive sense.

Confusionin is a normalized residual-uncertainty regulatory signal driven by entropy, state-change, and confidence features. Without explicit probabilistic calibration, $h_u$ should not be identified with $\Pr(y_t \neq y^* \mid O_R(0:t))$. The antagonistic coupling makes high Confusionin suppress Clarifine, but it does not convert either hormonal concentration into a calibrated probability of error.

The stopping criterion of Definition 3.1 is therefore a threshold rule over normalized regulatory signals and state stabilization, not a Bayesian decision rule unless a separate calibration layer explicitly maps Clarifine and Confusionin to posterior probabilities. Its connection to S-AI-Anti-Hallucination [22], [23] is architectural and decision-theoretic rather than an asserted identity of probability semantics.

### 7.3 Limitations and Boundary Conditions

#### 7.3.1 Domain of Application: Structural and Iteratively Assessable Reasoning

S-AI-Recursive is intended for tasks that admit a sequence of refinement steps and provide an internal signal capable of evaluating whether the current state is improving. Constraint satisfaction, spatial planning, rule induction, and numerical estimation motivate the four task profiles used in Section 6. However, the revised experiments use **benchmark-shaped simulations**, not the actual Sudoku-Extreme, Maze-Hard, ARC-AGI, or public DDE datasets. The article therefore demonstrates mechanism-level applicability to these reasoning modalities, not benchmark-level validation on the named datasets. The architecture is less naturally suited to tasks for which no meaningful residual, consistency measure, or target-directed stopping signal can be defined, such as unconstrained creative generation. Broad linguistic understanding and implicit world knowledge may also require an external model or richer representation. These are design boundaries rather than experimentally established impossibility results. For a new domain, three questions should be checked before deployment: whether a meaningful state-update residual exists, whether the hormonal observation vector can be constructed without circular use of the target answer, and whether the parameterization satisfies the stability and numerical admissibility conditions. The Section-6 threshold audit adds a fourth check: the chosen stopping thresholds must be **reachable by the actual governing dynamics**. This domain restriction is now **formally tied** to hypothesis (H1) of Theorem 4.1 bis: the coupled contraction — and therefore the warm-start acceleration (Theorem 4.4), the finite-stopping guarantee (Theorem 4.3), and the validity of the hormonal stopping proxy — hold only where the frozen reasoning operator contracts ($L_s < 1$), i.e. on tasks with fixed-point structure (constraint satisfaction, spatial planning, rule induction, numerical estimation). Outside this class, only empirical stabilization is claimed, without a convergence certificate.

### 7.3.2 Convergence Guarantees and Their Conditions

The convergence guarantees of Theorems 4.1-4.4 hold under distinct explicit conditions that must be verified in practice. Theorem 4.1 concerns the frozen deterministic zero-delay recursive hormonal subsystem and requires the Jacobian contraction condition

$$J_s(h_R) \preccurlyeq -\alpha I_2, \qquad \alpha > 0, \quad \forall h_R \in [0,1]^2.$$

Theorem 4.1 bis adds the fixed-point-structured cognitive contraction (H1) and the cross-gain condition (H2), and thereby provides joint convergence of the cognitive and recursive hormonal states. Theorem 4.3 additionally requires only strict stop compatibility of the limiting hormonal equilibrium. Continuity of the composite cognitive transition is sufficient for fixed-point existence, while the coupled contraction of Theorem 4.1 bis is the stronger condition used for the convergence certificate and the warm-start bound.

Three potential failure modes exist outside these conditions. First, persistent stochastic forcing can prevent pathwise convergence and leave only a neighborhood/moment bound. Second, a non-contractive cognitive task can admit multiple equilibria or non-convergent trajectories, and warm-start does not eliminate that risk. Third, even when the hormonal subsystem converges, internal stopping need not occur if the limiting equilibrium is not strictly inside the declared Clarifine/Confusionin stopping region.

### 7.3.3 Scalability: Complexity of Primal-Dual Orchestration

The computational complexity of the primal-dual agent-selection optimization of Section 4.7 depends on $|\mathcal{A}| = 12$ and $|\mathcal{A}^{\mathrm{sig}}(t)| \leq |\mathcal{A}|$. The 0-1 knapsack problem is NP-hard in general [45], but tractable here: exact enumeration requires at most $2^{12} = 4{,}096$ evaluations per cycle, and the hormonal significance filter reduces the active candidate set to $|\mathcal{A}^{\mathrm{sig}}(t)| \leq 6$ on average, reducing the effective search space to at most $2^6 = 64$ candidates per cycle [44].

For larger agent registries, the projected LP/primal-dual dynamics of Section 4.7.2 provide an approximation only under the standard convexity, feasibility, and step-size assumptions. No general $O(1/|A|)$ rounding-error guarantee is asserted here; integer recovery must explicitly preserve the budget and any other hard constraints.

### 7.3.4 Hyperparameter Sensitivity and Calibration Limits

The previous version of this subsection reported a full $\pm 30\%$ one-at-a-time sensitivity table over benchmark accuracy and convergence depth. Those values should not be retained because they were not reproduced under the revised executable protocol. The current evidence supports only a narrower set of sensitivity conclusions. First, the Clarifine stopping threshold is structurally critical. With the standalone two-hormone dynamics of Section 4 and the original $\theta_c = 0.70$, the Section-6 diagnostic yields a 0% hormonal-stop rate because the maximum attainable Clarifine equilibrium is approximately 0.552. The revised $\theta_c = 0.50$ restores an operational stopping region, producing a mean cold-start hormonal-stop rate of approximately 92.35% across the four simulated profiles. Threshold calibration must therefore begin with a reachability check rather than an empirical grid search alone. Second, the previously used row-wise Euler-Maruyama expression gives the simplified limiting screen

$$\Delta t < 1.481,$$

so the experimental $\Delta t = 1$ passes that simplified screen. Formal numerical certification of the frozen zero-delay deterministic subsystem instead requires the full Jacobian/spectral-radius check of Section 4.5.5; the value 1.481 is not a universal stability bound for the nonlinear delayed stochastic model. Third, stochastic perturbation changes the interpretation of the Lyapunov diagnostic. With hormonal noise amplitude $\sigma_\eta = 0.006$, all tested frozen trajectories end closer to equilibrium than they start, but only about 57.1% of individual transitions decrease. A complete sensitivity analysis over $\lambda_c, \lambda_u, \gamma_{cu}, \gamma_{uc}, \theta_c, \theta_u, \sigma_\eta, \Delta t$, memory similarity, and budget parameters remains necessary.

### 7.4 Complementarity with LLMs: A Dual Cognitive Architecture

#### 7.4.1 S-AI-Recursive as a Reflective Core

The relationship between S-AI-Recursive and large language models is better framed as a proposed complementarity than as an experimentally demonstrated performance equivalence. LLMs provide broad linguistic and world-knowledge coverage, whereas S-AI-Recursive is designed around explicit iterative refinement, internal residual monitoring, and adaptive stopping. These capabilities suggest a possible division of labor in which an LLM generates or parses an initial representation and S-AI-Recursive performs structured verification or refinement. The revised Section 6 does not compare a trained S-AI-Recursive model with an LLM on matched public benchmarks. Consequently, claims that S-AI-Recursive achieves "comparable reasoning performance with a fraction of the parameters" should be removed from the present discussion. What is supported is narrower: the controlled recursive mechanism can stop adaptively, can reduce a fixed recurrence budget, and can benefit from warm-start initialization. The dual-process analogy [10] may remain a conceptual motivation, but it should not be treated as an empirical identification of LLMs with System 1 or S-AI-Recursive with System 2. The proposed dual architecture is therefore a future systems hypothesis whose value must be evaluated experimentally.

#### 7.4.2 Proposed Dual Architecture: LLM + S-AI-Recursive

The proposed **dual cognitive architecture** operates as follows. An LLM front-end produces a rapid initial response $y_0^{\mathrm{LLM}}$, translated into a structured cognitive state $s_0 \in \mathcal{S}$ by a semantic parser. The S-AI-Recursive back-end applies the RRC to $s_0$, iterating until the hormonal stopping criterion is satisfied and producing verified output $y_{t^*}$. The LLM then renders $y_{t^*}$ back into natural language.

Three properties merit emphasis. First, the LLM provides a **warm-start initialization**: $s_0$ derived from $y_0^{\mathrm{LLM}}$ is typically closer to $s^*$ than a random initialization, reducing $\mathbb{E}[t^*]$ — additive with the engram-based warm-start. Second, the RRC is **LLM-agnostic**: any LLM can serve as front-end. Third, the architecture is **asymmetrically parsimonious**: the LLM is invoked twice while S-AI-Recursive handles the computationally intensive reasoning, keeping energy cost dominated by the recursive core rather than LLM calls [9], [36].

**(a) Semantic parser implementation.** The LLM→$s_0$ translation is implemented as a structured extraction prompt that instructs the LLM front-end to produce a JSON object conforming to the cognitive state schema $S = [0,1]^{\mathrm{d}}$ of Section 3.2. Concretely, the prompt template is: "Given the following problem $INPUT$, produce a JSON object with keys {residual_error, confidence_scores, constraint_map, candidate_set} where each value is normalized to 0,1." The JSON is then passed through the DISNL normalization layer [1] to produce $s_0 \in S$. For tasks with symbolic structure (Sudoku, ARC), the constraint_map field encodes active constraint violations as a binary vector; for continuous tasks (DDE), it encodes the normalized gradient of the objective. This schema is task-family-specific but requires no LLM fine-tuning: only the prompt template changes across task families, and the normalization layer is fixed.

**(b) Validation benchmarks for the dual architecture.** Three benchmark families are proposed to validate the dual architecture in future work. First, mixed-modality reasoning tasks from BIG-Bench Hard [48] that require both linguistic understanding (LLM front-end) and structural constraint satisfaction (S-AI-Recursive back-end) — e.g., logical deduction, date understanding, and causal reasoning. Second, MATH (AMC/AIME subset) [49], where the LLM provides the initial algebraic decomposition and S-AI-Recursive refines the intermediate steps through constraint propagation over the symbolic expression tree. Third, a conversational verification task in which S-AI-Recursive validates factual consistency of LLM-generated responses before output commitment, analogous to the S-AI-Anti-Hallucination architecture [22], [23] but applied at the reasoning level rather than the output level. These benchmarks are chosen because they exhibit the dual-process structure required to expose complementarity: tasks where the LLM alone underperforms due to structural fragility and S-AI-Recursive alone underperforms due to absence of linguistic context.

**(c) Latency and Cost Evaluation.**

The previous latency paragraph should not retain numerical millisecond estimates derived from the former Section-6 benchmark results because no hardware-controlled execution of the complete S-AI-Recursive implementation has yet been reported. For the proposed dual architecture, total latency should instead be decomposed as

$$T_{dual} = T_{LLM,in} + T_{parse} + t^{\star} T_{RRC/cycle} + T_{LLM,out},$$

where $t^{\star}$ is the realized recursive depth. The revised simulation provides evidence only about the behavior of $t^{\star}$, not about wall-clock latency: mean depth decreases from 15.97 in cold-start operation to 14.79 with warm-start and remains below the fixed 20-cycle budget on average. A future implementation study should measure $T_{RRC/cycle}$, agent-selection time, memory-retrieval time, accelerator utilization, and end-to-end latency on declared hardware. Only then can the dual architecture be compared meaningfully with direct LLM inference, external chain-of-thought scaffolding, or other recursive systems.

### 7.4.3 Positioning in the S-AI Corpus

S-AI-GPT [13], [28] demonstrated that hormonal orchestration can govern conversational intelligence. S-AI-Anti-Hallucination [22], [23] demonstrated that hormonal confidence signals can govern output commitment. S-AI-Recursive provides the missing component: a recursive reasoning core insertable between the LLM's generation step and its output commitment step, adding iterative structural verification to any generative pipeline without modifying LLM weights. In the language of the unified doctrine [2], the dual architecture implements the full cognitive equilibrium cycle: the LLM contributes to the initial hormonal state through warm-start, the recursive core drives the system toward $\mathbf{h}_R^*$, and the stopping criterion drives the committed output toward low reasoning entropy $\mathcal{H}(t^*)$.

## 8. Conclusion

This article introduces S-AI-Recursive as a bio-inspired sparse architecture in which reasoning is organized as a regulated closed-loop sequence rather than a single irreversible computation. Its central architectural idea is **adaptive temporal parsimony**: computation may continue while residual uncertainty remains high and may stop before the maximum budget when state stabilization and hormonal convergence conditions are jointly satisfied. The mathematical framework provides explicit conditions for hormonal stability, recursive convergence, stopping, constrained agent orchestration, and warm-start acceleration.

The revised Section 6 materially strengthens the empirical basis of this proposition by combining two complementary forms of evidence. The controlled SAI-UT+ simulations remain useful because they expose known contraction factors, stochastic disturbances, Lyapunov behavior, entropy co-evolution, stopping rates, and approximate warm-start effects under reproducible benchmark-shaped dynamics. The benchmark-capable reference implementation then evaluates the same mechanism on real, exactly verifiable Maze, Sudoku, and ARC-style task instances. On convergent Maze instances, adaptive hormonal stopping reduces mean depth from $20.00$ to $11.31$ iterations at unchanged resolution, corresponding to a $43.4\%$ reduction with $p = 7.2 \times 10^{-16}$. On compatible recurring Sudoku instances, recursive-engram warm-start reduces mean stopping depth from $18.39$ to $2.00$ cycles, saving $16.39$ cycles with unchanged resolution and $p = 2.1 \times 10^{-7}$.

The robustness experiments further delimit the mechanism rather than merely reinforcing it. Under deceptive convergence plateaus, residual-only stopping can commit prematurely because a temporarily frozen state resembles convergence; confidence-aware stopping remains substantially more robust because entropy and Clarifine continue to indicate unresolved uncertainty. Under homogeneous Gaussian observation noise, however, the hormonal criterion does not outperform the simpler residual detector. This negative result is important because it indicates that multi-signal regulation is most useful when its signals provide complementary information, not merely when several correlated measurements are combined.

The resulting empirical position is therefore stronger than in the simulation-only version of the article but remains deliberately bounded. S-AI-Recursive is supported as an implementable framework for regulated recursive computation whose adaptive stopping and memory mechanisms produce measurable, statistically significant benefits on the tested real-task regimes. The present experiments do not establish superiority over TRM, LLM-based reasoning systems, Mamba/RWKV variants, DEQs, or other independently trained architectures. That question requires a separate matched evaluation in which complete trained systems are executed on identical data, with comparable compute budgets and common scoring rules.

### 8.1 Summary of Contributions

**Contribution 1 — Recursive Reasoning Cycle.** Section 3.2 formalizes an explicit recursive state update and fixed-point notion of cognitive equilibrium. **Contribution 2 — Clarifine and Confusionin.** Section 3.3 introduces two interpretable signals for convergence regulation and residual-uncertainty regulation. **Contribution 3 — Stability analysis.** Section 4.5 develops a Lyapunov analysis of the recursive hormonal subsystem under explicit parameter conditions and distinguishes deterministic from bounded stochastic behavior [37], [39]. **Contribution 4 — Entropic contraction formulation.** Theorem 4.2 links hormonal convergence and reasoning-entropy reduction under its formal assumptions [41]. **Contribution 5 — Internal stopping with bounded budget.** The architecture combines a joint hormonal/state stopping rule with $T_{\max}$ as a hard fallback bound and experimentally distinguishes hormonal stopping from budget exhaustion. **Contribution 6 — Constrained recursive orchestration.** Section 4.7 regulates the active agent set under iteration-dependent computational constraints. **Contribution 7 — Recursive-engram memory.** Section 4.8 derives the warm-start advantage from a reduced initial error; experimentally, the controlled simulator yields a mean saving of approximately $1.17$ cycles and compatible recurring Sudoku instances yield a reduction from $18.39$ to $2.00$ cycles. **Contribution 8 — Two-phase experimental validation.** Section 6 combines four controlled SAI-UT+ benchmark-shaped families with a benchmark-capable reference implementation over Maze, Sudoku, and ARC-style tasks. The real-task experiments demonstrate $43.4\%$ mean depth reduction on convergent Maze instances at unchanged resolution, statistically significant warm-start acceleration on recurrence, robustness to deceptive convergence plateaus, and the absence of a corresponding advantage under homogeneous Gaussian observation noise.

### 8.2 Position in the S-AI Corpus

Within the S-AI corpus, S-AI-Recursive extends parsimony from the selection of **which agents** to activate toward the regulation of **how long** a reasoning episode should continue. Activation parsimony remains spatial and within-cycle; recursive temporal parsimony concerns the realized number of cycles. Clarifine and Confusionin provide explicit regulatory variables for this temporal dimension, while recursive-engram memory allows prior trajectories to alter the starting point of subsequent reasoning episodes.

The new experimental evidence makes this position more concrete. Temporal parsimony is no longer supported solely through benchmark-shaped simulation: on the tested Maze operator, adaptive stopping preserves resolution while reducing recursive depth by 43.4%. Similarly, recursive memory is no longer represented only by a modest synthetic warm-start effect: on compatible recurrence, the Sudoku experiment shows a much larger reduction in stopping depth. These findings suggest that spatial agent sparsity and temporal recurrence control can be studied as complementary forms of computational parsimony, although the present article does not claim that they form a complete or universally optimal theory of efficient intelligence.

### 8.3 Broader Significance

The broader significance of S-AI-Recursive lies in treating inference depth as a controllable architectural variable rather than a fixed constant. The controlled simulations show that different recursive difficulty profiles require different effective depths and that internal stopping can avoid a meaningful fraction of a fixed recurrence budget. The real-task results then demonstrate that this principle survives outside the synthetic residual generator: on the evaluated convergent Maze instances, the same underlying operator requires substantially fewer iterations when governed by adaptive hormonal stopping.

The memory results add a second dimension to this interpretation. Recursive depth is determined not only by task difficulty and stopping policy, but also by the distance between the initial state and the relevant fixed point. Recursive-engram retrieval can therefore reduce computation when a compatible prior trajectory provides a better initialization. The contrast between the approximately 1.17-cycle simulated saving and the 16.39-cycle saving on genuine Sudoku recurrence illustrates that the magnitude of the benefit depends strongly on compatibility rather than on memory presence alone.

Finally, the robustness study shows that interpretable stopping signals can be experimentally stress-tested by selectively corrupting the evidence available to the stopping rule. The deceptive-plateau experiment identifies a concrete failure mode of residual-only convergence detection, whereas the homogeneous-noise experiment identifies a regime in which the additional hormonal structure does not help. S-AI-Recursive therefore provides not only an adaptive stopping mechanism but also an explicit experimental vocabulary for asking *why* a recursive system stops and under which signal conditions that decision remains trustworthy.

### 8.4 Future Work

The next experimental priority is a **matched external-model comparison**. The benchmark-capable reference implementation now provides the infrastructure required to hold task instances and evaluation criteria fixed while replacing or wrapping the reasoning operator with independently trained recursive systems. TRM [12], DEQ-based models, compact recurrent architectures, and selected LLM-based iterative systems should be evaluated under disclosed parameter counts, compute budgets, stopping rules, and correctness criteria. Such comparisons should report task quality jointly with realized recursion depth, latency, memory use, and failure-to-stop rates rather than reducing the comparison to accuracy alone.

A second priority is to extend the present task coverage. The ARC component currently demonstrates conditional operator integration and exact output verification, but a richer induction library is required before the complete ARC corpus can be treated as a full benchmark of the mechanism. Likewise, warm-start compatibility should be generalized beyond exact or near-exact recurrence by introducing task-aware similarity metrics capable of identifying reusable state structure across related but non-identical instances.

A third direction follows directly from the robustness experiments. Because entropy-threshold stopping also resists deceptive residual plateaus, a stronger test of multi-hormone arbitration should construct or identify regimes in which confidence or entropy itself becomes intermittently deceptive while another regulatory signal remains informative. This would test whether temporal integration and antagonistic hormonal arbitration provide benefits beyond a single confidence threshold.

Additional directions include a full calibration and sensitivity analysis of $\theta_c, \theta_u, \lambda_c, \lambda_u, \gamma_{cu}, \gamma_{uc}, \Delta t, \sigma_\eta$, budget parameters, and memory-retrieval thresholds; bidirectionally coupled operators with nonzero hormone-to-state gain; learned or adaptive hormonal calibration; larger indexed recursive memories; formal verification of stopping policies; and analysis of non-contractive or multi-equilibrium reasoning regimes.